# Sensitivity study of ANFIS model parameters to predict the pressure gradient with combined input and outputs hydrodynamics parameters in the bubble column reactor

Shahaboddin Shamshirband [1,2], Amir Mosavi [3,4] *, Kwok-wing Chau [5]

[1] Department for Management of Science and Technology Development, Ton Duc Thang University, Ho Chi Minh City, Vietnam

[2] Faculty of Information Technology, Ton Duc Thang University, Ho Chi Minh City, Vietnam: Shahaboddin.shamshirband@tdtu.edu.vn

[3] Institute of Automation, Kando Kalman Faculty of Electrical Engineering, Obuda University, Budapest-1034, Hungary; amir.mosavi@kvk.uni-obuda.hu

[4] School of the Built Environment, Oxford Brookes University, Oxford OX3 0BP, UK; a.mosavi@brookes.ac.uk

[5] Department of Civil and Environmental Engineering, Hong Kong Polytechnic University, Hong Kong, People's Republic of China; dr.kwok-wing.chau@polyu.edu.hk

## Abstract

Intelligent algorithms are recently used in the optimization process in chemical engineering and application of multiphase flows such as bubbling flow. This overview of modeling can be a great replacement with complex numerical methods or very time-consuming and disruptive measurement experimental process. In this study, we develop the adaptive network-based fuzzy inference system (ANFIS) method for mapping inputs and outputs together and understand the behavior of the fluid flow from other output parameters of the bubble column reactor. Neural cells can fully learn the process in their memory and after the training stage, the fuzzy structure predicts the multiphase flow data. Four inputs such as x coordinate, y coordinate, z coordinate, and air superficial velocity and one output such as pressure gradient are considered in the learning process of the ANFIS method. During the learning process, the different number of the membership function, type of membership functions and the number of inputs are examined to achieve the intelligent algorithm with high accuracy. The results show that as the number of inputs increases the accuracy of the ANFIS method rises up to $R^2 > 0.99$ almost for all cases, while the increment in the number of rules has a effect on the intelligence of artificial algorithm. This finding shows that the density of neural objects or higher input parameters enables the moded for better understanding. We also proposed a new evaluation of data in the bubble column reactor by mapping inputs and outputs and shuffle all parameters together to understand the behaviour of the multiphase flow as a function of either inputs or outputs. This new process of mapping inputs and outputs data provides a framework to fully understand the flow in the fluid domain in a short time of fuzzy structure calculation.

## 1. Introduction

The extensive application of bubble column reactors is indebted to several benefits provided both by their engineering and functioning in comparison to other reactors. Firstly, their outstanding mass transfer and heat features mean mass transfer and high heat coefficients. Slight upkeep and less operation expenses are needed because of compression and lacking moving parts. The catalyst and/or other packing material are highly durable (Degaleesan, Dudukovic, & Pan, 2001). Furthermore, other vantages include withdrawal ability, online catalyst addition, and plug-free operation rendering bubble columns a popular reactor option (Prakash, Margaritis, Li, & Bergougnou, 2001). Within the last few decades, research on key hydrodynamic and functioning parameters distinguishing their function have has substantially focused on their industrial standing, widespread applicability, and designing and scaling up bubble column reactors. Recently, studies on bubble columns have often addressed such subjects as gas holdup (Anabtawi, Abu-Eishah, Hilal, & Nabhan, 2003; Bouaifi, Hebrard, Bastoul, & Roustan, 2001; Forret, Schweitzer, Gauthier, Krishna, & Schweich, 2003; Luo, Lee, Lau, Yang, & Fan, 1999; Shimizu, Takada, Minekawa, & Kawase, 2000; Tang & Heindel, 2004; Veera, Kataria, & Joshi, 2004; S. Wang et al., 2003), bubble characteristics (Essadki, Nikov, & Delmas, 1997; Lapin, Paaschen, Junghans, & Lübbert, 2002; H Li & Prakash, 1999, 2000; Prakash et al., 2001; Schäfer, Merten, & Eigenberger, 2002), flow regimes and computational fluid dynamics (Buwa & Ranade, 2002; Degaleesan et al., 2001; M. Dhotre, Ekambara, & Joshi, 2004; Michele & Hempel, 2002; Ruzicka, Zahradnık, Drahoš, & Thomas, 2001; Thorat & Joshi, 2004; Mosavi et al. 2019), local and mean heat transfer determinations (Chen, Hasegawa, Tsutsumi, Otawara, & Shigaki, 2003; Cho, Woo, Kang, & Kim, 2002; Hanning Li & Prakash, 2001; H Li & Prakash, 2002; Lin & Wang, 2001), and mass transfer (Behkish, Men, Inga, & Morsi, 2002; Krishna & Van Baten, 2003; Maalej, Benadda, & Otterbein, 2003; Vandu & Krishna, 2004; Verma & Rai, 2003). Common topics examined in such investigations include the impacts of column internals design, column dimensions, working settings (temperature and i.e. pressure), solid type and influence of superficial gas velocity, and concentration. A number of empirical examinations have focused on the evaluation of operational settings, column dimensions and slurry physical properties affecting the function of bubble columns (Pino et al., 1992). The fluid dynamic determination of bubble column reactors significantly influences the functioning and performance of bubble columns. As reported in previous researches, empirical outcomes gained by parameter surveys are strongly dependent upon the predominant regime near the gas distributors, middle of the rector or near the surface of the reactor (Kumar, Degaleesan, Laddha, & Hoelscher, 1976; Lefebvre & Guy, 1999; Rabha, Schubert, & Hampel, 2013; Şal, Gül, & Özdemir, 2013; Shah, Kelkar, Godbole, & Deckwer, 1982; A Sokolichin & Eigenberger, 1994). The flow regime in the reactor changes with different operating conditions such as superficial gas velocity, column dimensions, sparger arrangement and specifications, gas and liquid properties and reactor operating pressure and temperature (Buwa & Ranade, 2003; Deen, Solberg, & Hjertager, 2000; M. T. Dhotre, Niceno, Smith, & Simiano, 2009; Díaz et al., 2008; Masood & Delgado, 2014; Pourtousi, Ganesan, Kazemzadeh, Sandaran, & Sahu, 2015; Pourtousi, Ganesan, Sandaran, & Sahu, 2016). There are two main flow regimes such as homogeneous (low rate of break-up and coalescence) and heterogeneous regime (high rate of break-up and coalescence) (Masood, Khalid, & Delgado, 2015; Pourtousi, Ganesan, & Sahu, 2015; Silva, d'Ávila, & Mori, 2012; Sobrino, Acosta-Iborra, Izquierdo-Barrientos, & De Vega, 2015). The homogeneous flow regime appears at very small inlet velocity roughly below 5 cm/s with a low rate of merging bubbles. In this condition, the turbulence and shear flow in the reactor is not large enough to break fluids or merge the interface (Fan, 1989; Hills, 1974; T. Ziegenhein, Rzehak, & Lucas, 2015). The characteristics of this flow regime are bubbles of rather uniform low dimensions and high velocities (Lopez de Bertodano, Lahey Jr, & Jones, 1994; Schumpe & Grund, 1986). An even bubble dissemination and rather mild mixing are observable above the whole cross-section of the column (Besagni, Guédon, & Inzoli, 2016; Clift, 1978; Hyndman, Larachi, & Guy, 1997; Joshi, 2001). In practice, there is not any break-up or bubble coalescence, so in this regime, bubble size is nearly entirely determined by the system properties and sparger design (Burns, Frank, Hamill, & Shi, 2004; Rampure, Kulkarni, & Ranade, 2007; Thorat, Joshi, & Science, 2004; Xing, Wang, & Wang, 2013).

Additionally, numerical methods such as CFD algorithms and established mathematical models support empirical investigations for the enhanced description of the procedures occurring in a bubble column reactor (Pfleger & Becker, 2001; Pourtousi, 2016; Pourtousi, Sahu, & Ganesan, 2014; Rzehak & Krepper, 2013; Simonnet, Gentric, Olmos, & Midoux, 2008; Tabib, Roy, & Joshi, 2008; H. Wang et al., 2014). It has been demonstrated that Computational Fluid Dynamics (CFD) to be very practical in the simulation of polyphase flows in the BCR, in particular, because of the substantial developments in numerical procedures and calculating power throughout the last two decades(de Bertodana, 1992; de Bertodano, 1992; Jay Sanyal, Marchisio, Fox, & Dhanasekharan, 2005). CFD can be useful for parametric investigations after validating the model to diminish experimental expenses(C. Azwadi, Razzaghian, Pourtousi, & Safdari, 2013; Krishna, Urseanu, Van Baten, & Ellenberger, 1999; McClure, Aboudha, Kavanagh, Fletcher, & Barton, 2015; Jayanta Sanyal, Vásquez, Roy, & Dudukovic, 1999). As one of the most predominant analytical tools, CFD is used for the analysis of heat and mass flow processes due to its multiple uses and the accessibility of various robust, easy to use computer software applying CFD. CFD as a numerical process comes with such downsides as precision and stableness(Buwa, Deo, & Ranade, 2006; McClure, Kavanagh, Fletcher, & Barton, 2013, 2014; Simonnet, Gentric, Olmos, & Midoux, 2007). Whereas the stableness problem can typically be addressed through lesser time steps, the accuracy issue is improvable via better discretization order and mesh quality for the determining equations(Besbes, El Hajem, Aissia, Champagne, & Jay, 2015; Islam, Ganesan, & Cheng, 2015; Alexande Sokolichin, Eigenberger, & Lapin, 2004; Xiao, Yang, & Li, 2013; Thomas Ziegenhein, Rzehak, Krepper, & Lucas, 2013). Nevertheless, improving discretization order and/or mesh quality may exacerbate the solution process stableness(Liu & Hinrichsen, 2014; McClure, Norris, Kavanagh, Fletcher, & Barton, 2015; Xiao et al., 2013; T. Ziegenhein et al., 2015).

To analyze CFD, therefore, no universal measure exists to realize the best integration of discretization order, mesh quality and time step. Thus, some CFD analyses may become overpriced because of lengthy analysis time and increased computing price. Intelligent methods can be a great alternative to replace with CFD analysis and predict the flow pattern(Ekambara, Dhotre, & Joshi, 2005; Laborde-Boutet, Larachi, Dromard, Delsart, & Schweich, 2009; Van Baten, Ellenberger, & Krishna, 2003). These methods due to having inexpensive algorithms structure, they can mimic the CFD results and enable researchers to simulate many CFD cases. Pourtousi et al. (M. Pourtousi, J. N. Sahu, P. Ganesan, S. Shamshirband, & G. Redzwan, 2015; Pourtousi, Zeinali, Ganesan, & Sahu, 2015) showed that soft computing methods can predict the microscopic and macroscopic fluid pattern with the condition of appropriate learning process. They used the ANFIS method for combination of CFD and smart modeling and this model showed the great ability in the prediction of CFD results. Intelligent algorithms provide a potent instrument to analyze intricate issues, including mass flow processes and heat, which maintains the quality of CFD method but necessitates lower CPU work than CFD(Ali Ghorbani, Kazempour, Chau, Shamshirband, & Taherei Ghazvinei, 2018; Chau, 2017; Chuntian & Chau, 2002; Moazenzadeh, Mohammadi, Shamshirband, & Chau, 2018; Wu & Chau, 2011; Yaseen, Sulaiman, Deo, & Chau, 2018). While creating machine learning tools by ANFIS, an important point is to choice appropriate parameters, such as the number of membership function (MF) and the type of membership function(Babanezhad, Rezakazemi, Hajilary, & Shirazian; Shamshirband, Babanezhad, & Mosavi, 2019). Selection of suitable factors for the learning procedure, including the presence of data P, is also of importance to represent the percentage of data existing in the training process. Previous research has focused on the selection of parameters and their effects on the ANFIS (Fritzke, 1997; J.-S. R. Jang, Sun, & Mizutani, 1997; Taha, Noureldin, & El-Sheimy, 2001). About specific training and testing datasets, the impacts of these parameters are analyzed on the final ANFIS performance, namely the training and testing processes. A sensitivity investigation of different tuning parameters of the machine learning method for better prediction of the process has not fully studied (Mosavi et al. 2019; Vargas et al. 2017; Mosavi et al. 2017). Additionally, after the learning process in the ANFIS structure, there are many possibilities to learn about the flow in the domain by creating a map between inputs and outputs. Therefore, in this study, we examined different tuning parameters of the ANFIS method to achieve the high accuracy of this method in the prediction. Additionally, we combined all outputs and inputs parameters together and study the effect of different inputs and outputs on the behavior of the multiphase flow.

## 2. Method

In order to stimulate the liquid and gas interactions, the two-stage model was applied, which was based on the Eulerian-Eulerian approach. The stages are treated like a continuum in the domain being considered in this approach. The basis of the Eulerian modeling framework is the momentum transport and ensemble-averaged mass equations.

Continuity equation:

$$\frac{\partial}{\partial t}(\rho_k \epsilon_k) + \nabla(\rho_k \epsilon_k u_k) = 0 \tag{1}$$

Momentum transfer equation:

$$\frac{\partial}{\partial t}(\rho_k \epsilon_k u_k) + \nabla(\rho_k \epsilon_k u_k u_k) = -\nabla(\epsilon_k \tau_k) - \epsilon_k \nabla p + \epsilon_k \rho_k g + M_{I,k} \tag{2}$$

The overall interfacial force that is active between the two stages is on the basis of the turbulent dispersion force and interphase drag force. These can be restated as Relation (3):

$$M_{I,L} = -M_{I,G} = M_{D,L} + M_{TD,L} \tag{3}$$

Tabib et al. provided a detailed description about the interfacial force models applied in the current work(Tabib et al., 2008).

In addition, the interfacial forces, the turbulence model is regarded as one of the main elements for capturing the bubble column's hydrodynamic characteristics. The k–ε model has been extensively applied over the last 20 years for describing the flow pattern that is observed in the bubble columns. This model is a low-cost model, which is sufficiently reliable because of low computational necessities and its simplicity. This turbulence model is used in the current study for all simulations. All parameters of the turbulence model are the same as the parameters described by Pourtousi et al(Pourtousi, 2012; M. Pourtousi, J. Sahu, P. Ganesan, S. Shamshirband, & G. Redzwan, 2015; M. Pourtousi, Mohammadjavad Zeinali, et al., 2015).

## 2.1. Geometrical structure

For the CFD study and generation of data-set, a cylindrical bubble column reactor is used with dimensions of 0.288 m (diameter) and 2.6 m (height). This reactor structure and the flow regime is same as the reactor used by Pfleger and Becker. The superficial velocity of the gas is 0.005 m/s at the room condition. Pfleger and Becker provide the details about the boundary conditions including outlet pressure and walls used in the current study. The inlet boundary condition is like the conditions used in Tabib et al(Tabib et al., 2008).

## 2.2. Grid

A structured grid is used in the domain that is based on the hexahedral grid. The grid type used is similar to the type utilized in Boutet et al(Laborde-Boutet et al., 2009).

## 2.3. ANFIS

For this study, the neural network is used to learn the data, while the fuzzy interface structure is implemented to predict the process. This process of learning and prediction within the framework of ANFIS structure is better than another individual model such as a single neural network or single fuzzy interface methods. Single core programming is used to learn and predict data. During the learning process, 70% of data is used for learning and the rest of the data is used for the evaluation process. This process of evaluation is applied in the computing code. The evaluation process is used for the validation of our code implementation. For the prediction process, we use the meshless method. In this case, the AI predicts the non-existing data in the CFD domain. Based on previous work, the number of inputs and percentage of training data have a large effect on the accuracy of training and testing.

ANFIS is defined as fuzzification and defuzzification framework for providing an exact prediction of the behavior in the nonlinear complicated systems.(Abdulshahed, Longstaff, & Fletcher, 2015; C. S. N. Azwadi, Zeinali, Safdari, & Kazemi, 2013; Choubin et al. 2019a; Choubin et al. 2019b; Riahi-Madvar et al. 2019; Rezakazemi et al. 2019; Dehghani et al. 2019; Mosavi, and Edalatifar 2018; ). The ANFIS structure used for prediction of the hydrodynamic properties in the 3D bubble column is shown in Figure 1. In the current work, (The x, y, and z coordinates, air superficial velocity) are used for obtaining (pressure gradient) as output. In the first layer, the inputs are classified in different function characteristics. As an example, the signal incoming $i^{th}$ rule function is stated as follows,

$$w_i = \mu_{Ai}(X)\, \mu_{Bi}(Y)\mu_{ci}(Z)\mu_{di}(\text{Vas}) \tag{4}$$

where $w_i$ denotes the signal out-coming from the node of the second layer and $\mu_{Ai}$, $\mu_{Bi}$ and $\mu_{Ci}$ denote the signals incoming from the MFs implemented on inputs, x coordination (X), y coordination (Y), z coordination (Z) and air superficial velocity (Vas), to the node of the second layer.

In the third layer, the relative value of each rule's firing strength is measured(J.-S. Jang, 1993; Varol, Koca, Oztop, & Avci, 2008). This value is equal to each layer's weight over the overall amount of firing strengths of all rules:

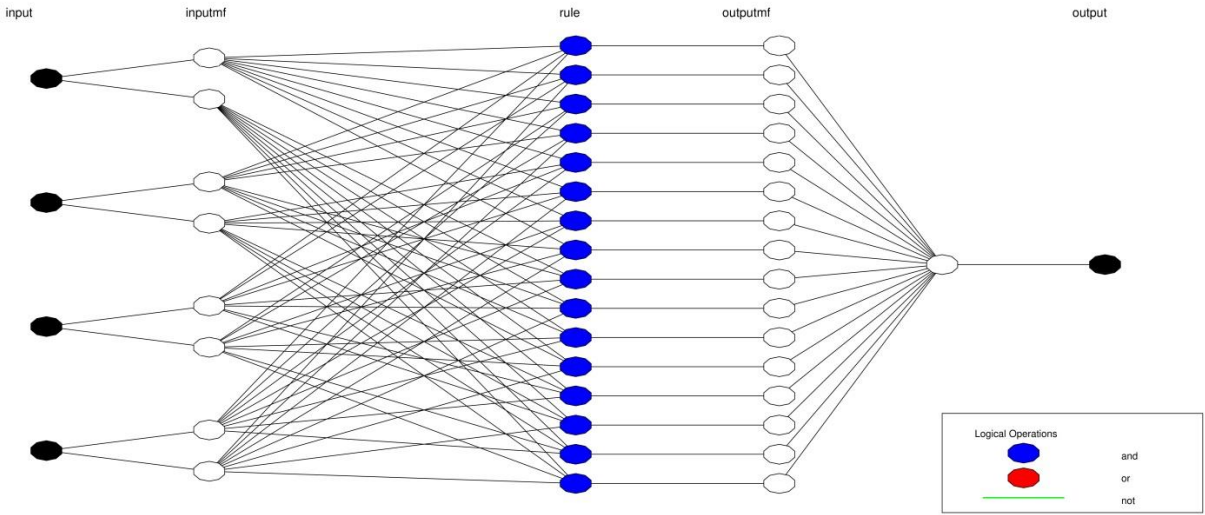

Figure1: Schematic of the ANFIS structure with four inputs.

$$\overline{w_i} = \frac{w_i}{\sum(w_i)} \tag{5}$$

Where $\overline{w_i}$ denotes the normalized firing strengths. In the fourth level integration algorithm, the calculation of a consequence if-then rule is applied that was suggested by previous study(Avila & Pacheco-Vega, 2009; Lei, He, Zi, & Hu, 2007; Ryoo, Dragojlovic, & Kaminski, 2005; Schurter & Roschke, 2000; Takagi & Sugeno, 1985).

Hence, the node function is stated as follows:

$$\overline{w_i} f_i = \overline{w_i}(p_i X + q_i Y + r_i Z + S_i V_{as} + t_i) \tag{6}$$

Where $p_i$, $q_i$, $r_i$, $s_i$ and $t_i$ denote the parameters' if-then rules, and they are known as the consequent parameters. The combination algorithms called a hybrid framework is used for updating all tuning parameters of the ANFIS method. Additionally, this algorithm, the gradient descent method updates the parameters of MFs, and the Least Square Estimate (LSE) method updates the consequent parameters.

## 3. Result and discussion

The x, y, and z coordinates, air superficial velocity, and pressure gradient are some of the parameters yielded by the CFD method. Due to the considerable computational time required for the CFD method, artificial intelligence (AI) can be highly helpful to obtain fluid characteristics at different points and dramatically reduce the computational time. This study intends to investigate the information obtained by the CFD method using AI algorithms (ANFIS method).

To commence the study using the ANFIS method, a part of the CFD output data was considered as the input, and the other part was used as the output. Accordingly, four inputs (viz., x coordinate, y coordinate, z coordinate, and air superficial velocity as input 1, 2, 3, and 4, respectively) and one output (viz., pressure gradient) were considered for the investigations. The number of iterations, the total size of data, and p-value (i.e., a percentage of the total data used in the training process) were considered 700, 6000, and 70%, respectively. In this study, 70% of the data was used in the training process, and 30% of the remaining data accompanied with 70% of the data related to the training process were scrutinized in the testing process.

The x coordinate and pressure gradient were initially considered as the input and output, respectively, to investigate the ANFIS intelligence. Considering 'number of membership = 2 functions' for all types of membership functions (MFs), namely generalized bell-shaped (gbellmf), Gaussian curve (gaussmf), Gaussian combination (gauss2mf), difference between two sigmoidal functions (dsigmf), product of two sigmoidal functions (psigmf), and triangular-shaped (trimf), training and testing processes were accomplished separately. As shown in Fig. 2 (a and b), under the best possible condition, $R^2$ equals 0.08, indicating that ANFIS lacks adequate intelligence hence further modifications are required to be applied to ANFIS method to promote its intelligence. By changing the number of MFs from two to four for all types of MFs, the learning process involving training and testing was separately evaluated. Results reveal little progress in the ANFIS intelligence improvement, and in the best condition possible, $R^2$ is equal to 0.11 (see Figs. 3a and 3b).

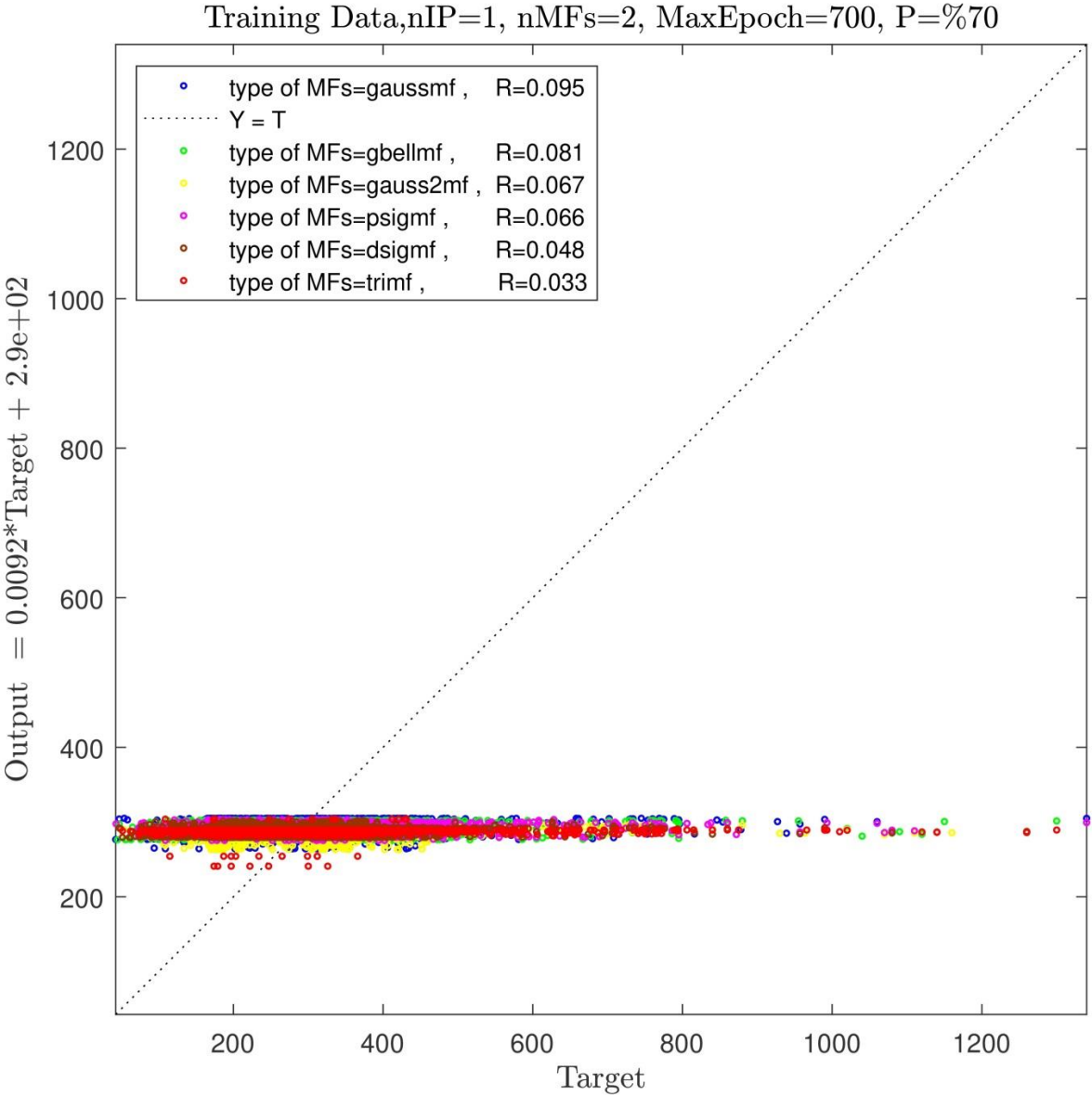

Figure 2(a): Training process, one input, number of MFs=2, various types of MFs.

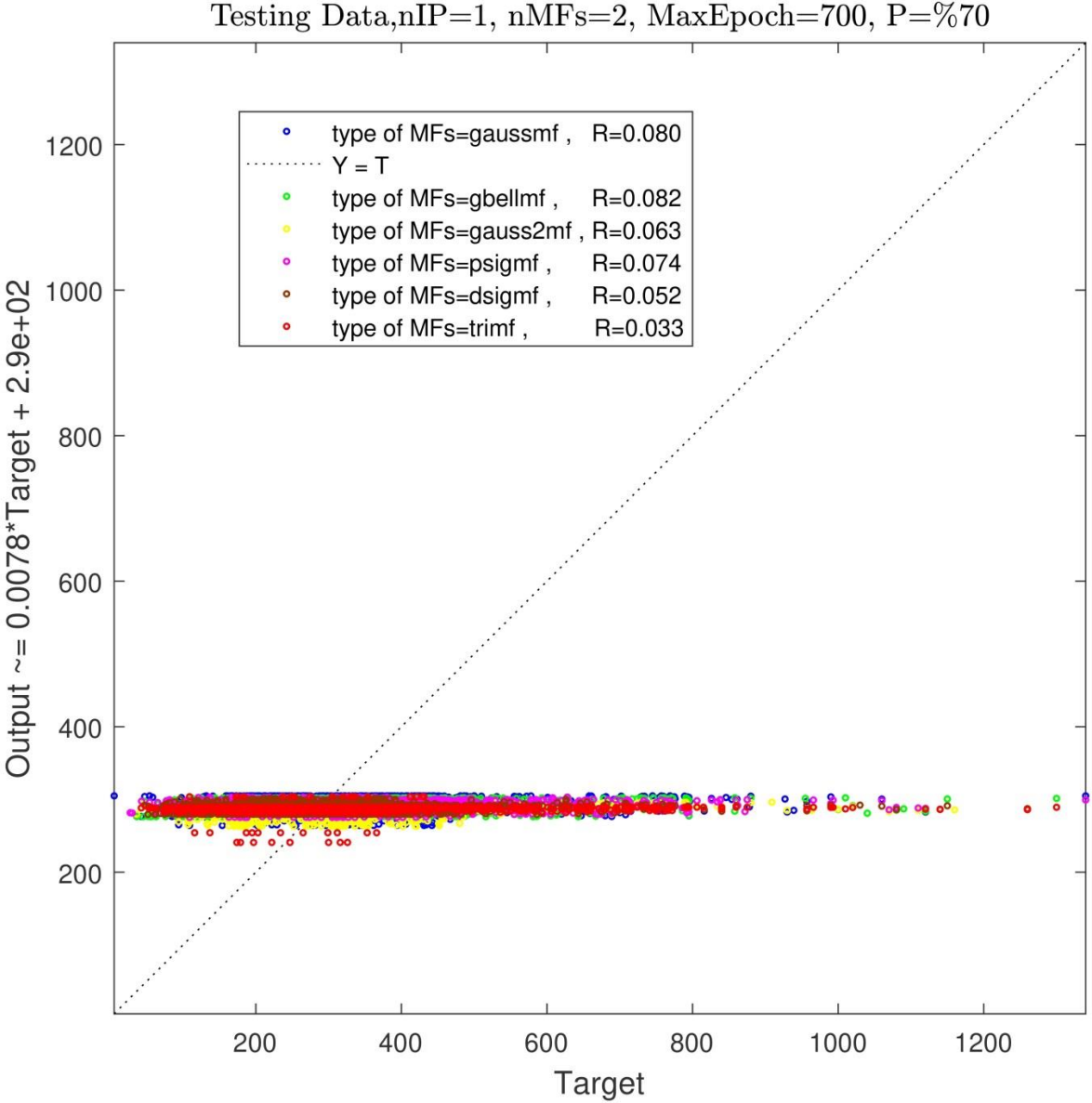

Figure 2(b): Testing process, one input, number of MFs=2, various types of MFs.

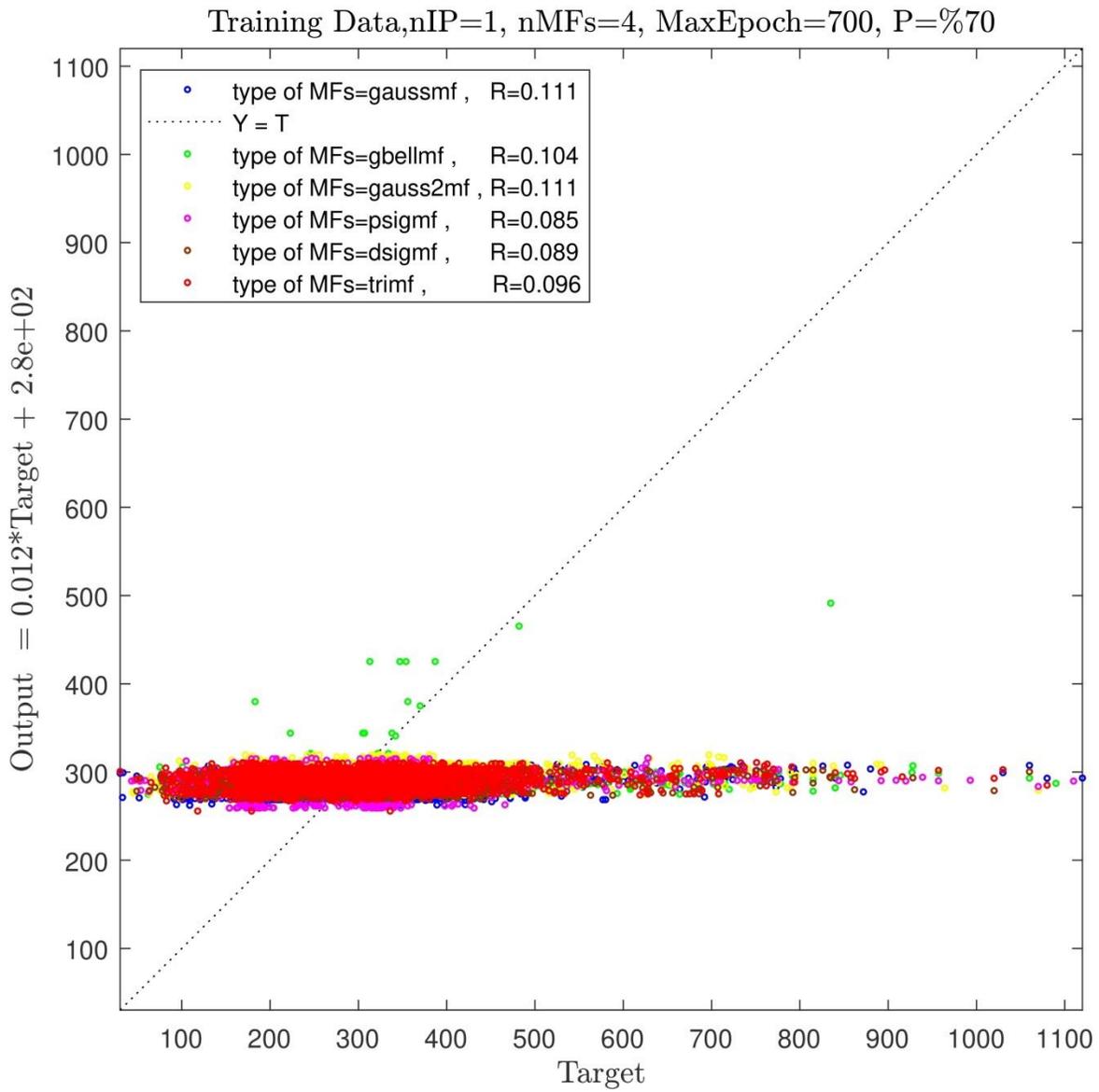

Figure 3(a): Training process, one input, number of MFs=4, various types of MFs.

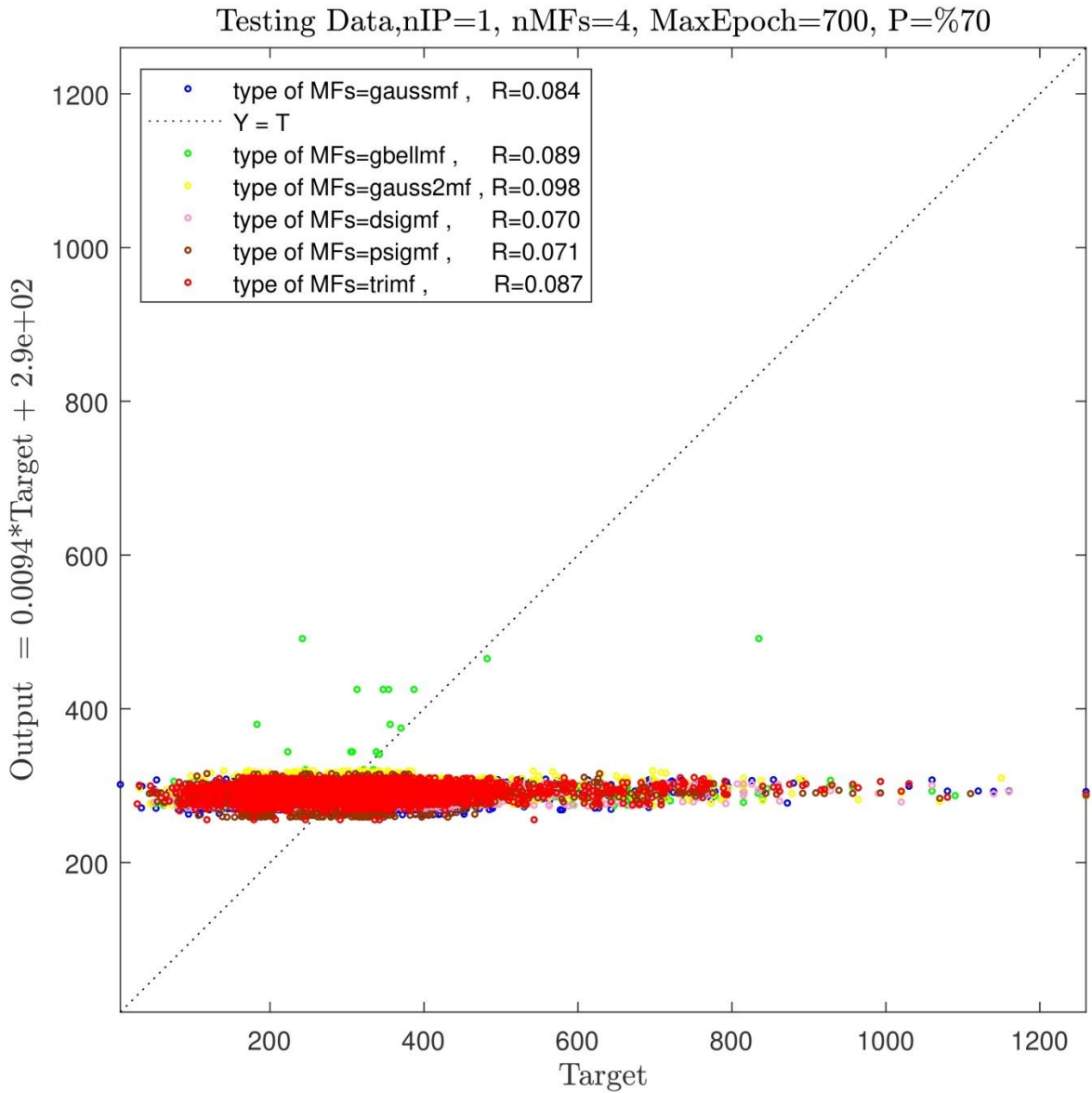

Figure 3(b): Testing process, one input, number of MFs=4, various types of MFs.

Accordingly, the number of MFs increased from four to six in all types of MFs, and then the ANFIS intelligence was evaluated as well. Figs. 4 (a) and 4 (b) illustrate no progress in ANFIS intelligence. Applying changes in the type and number of MFs, in the case that there is only one input, failed to make noticeable progress toward improving the intelligence; therefore, it was decided to increase the number of inputs from one to two for the rest of the investigations. To this end, x and y coordinates were considered the inputs, and pressure gradient was considered the output. The training and testing processes for all types of MFs (the number of MFs = 2) were separately analyzed using ANFIS method. The results of investigations, as observed in Figs. 5a and 5b, show an increase in the $R^2$ value from 0.12 to 0.64, representing that ANFIS intelligence increased by 52%. Such an increase, however, fails to suffice to reach full-fledged intelligence; thus, further investigations are needed. Therefore, by increasing the number of MFs to four, the change in the type of MFs was investigated.

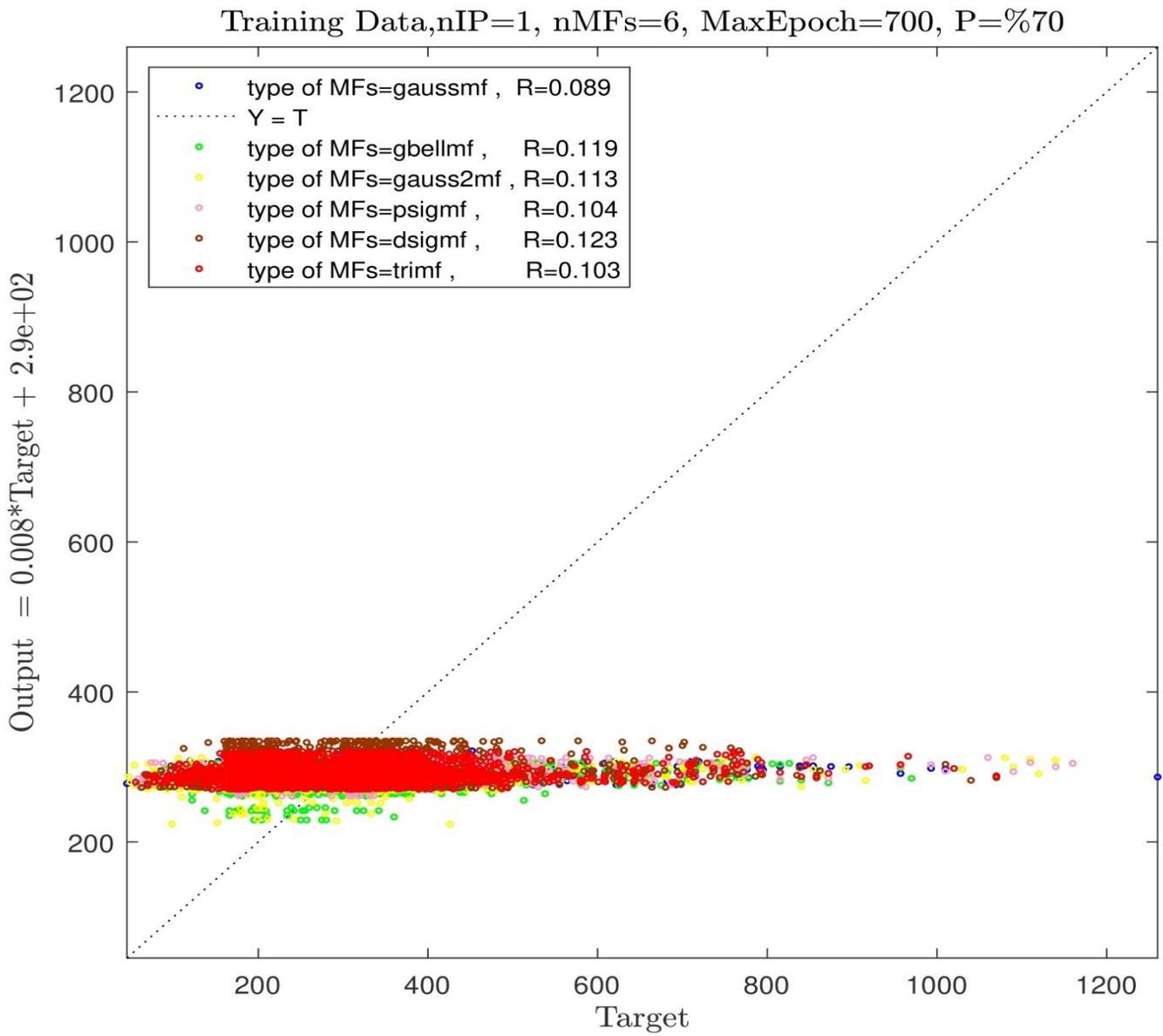

Figure 4(a): Training process, one input, number of MFs=6, various types of MFs.

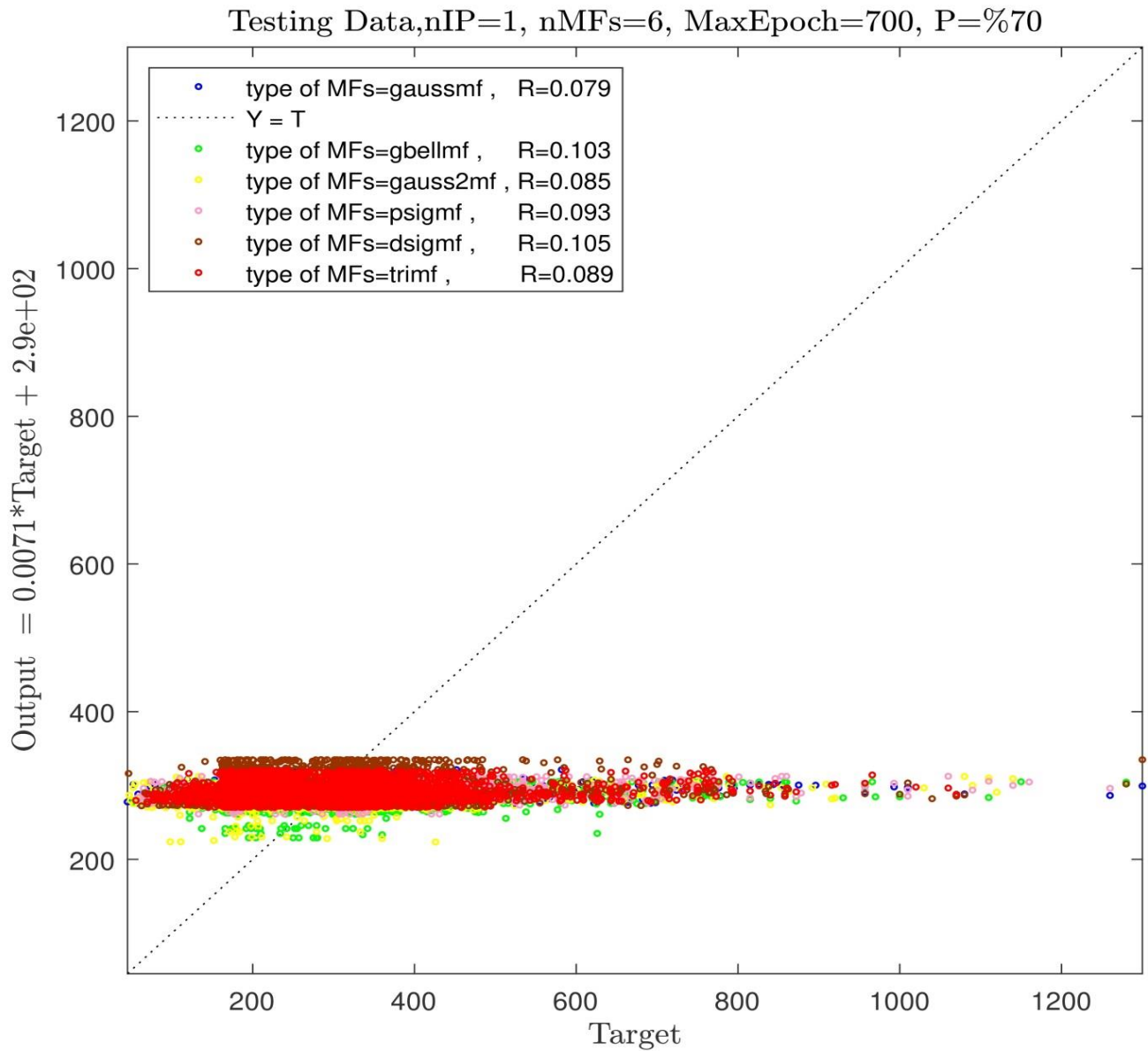

Figure 4(b): Testing process, one input, number of MFs=6, various types of MFs.

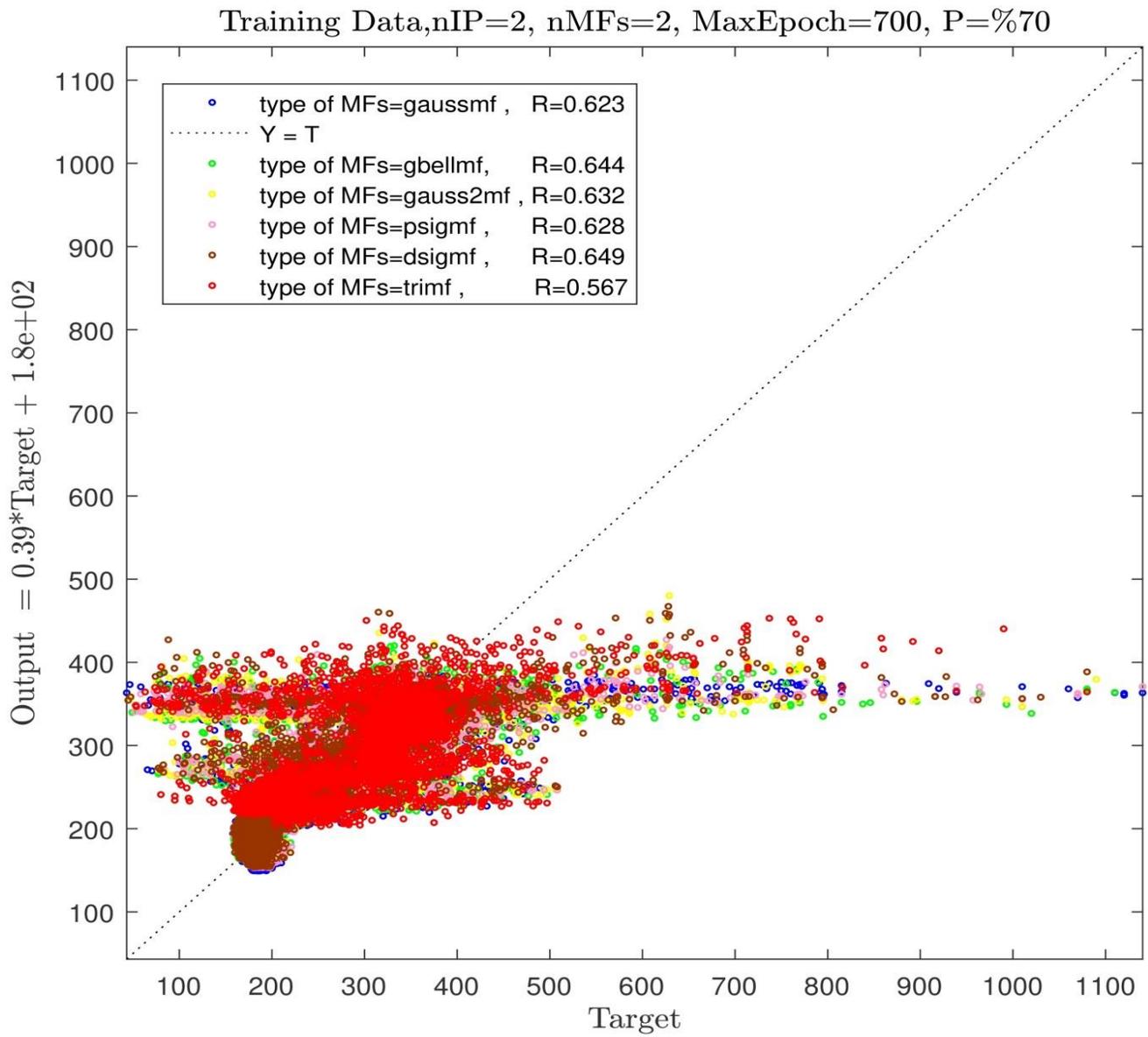

Figure 5(a): Training process, two inputs, number of MFs=2, various types of MFs.

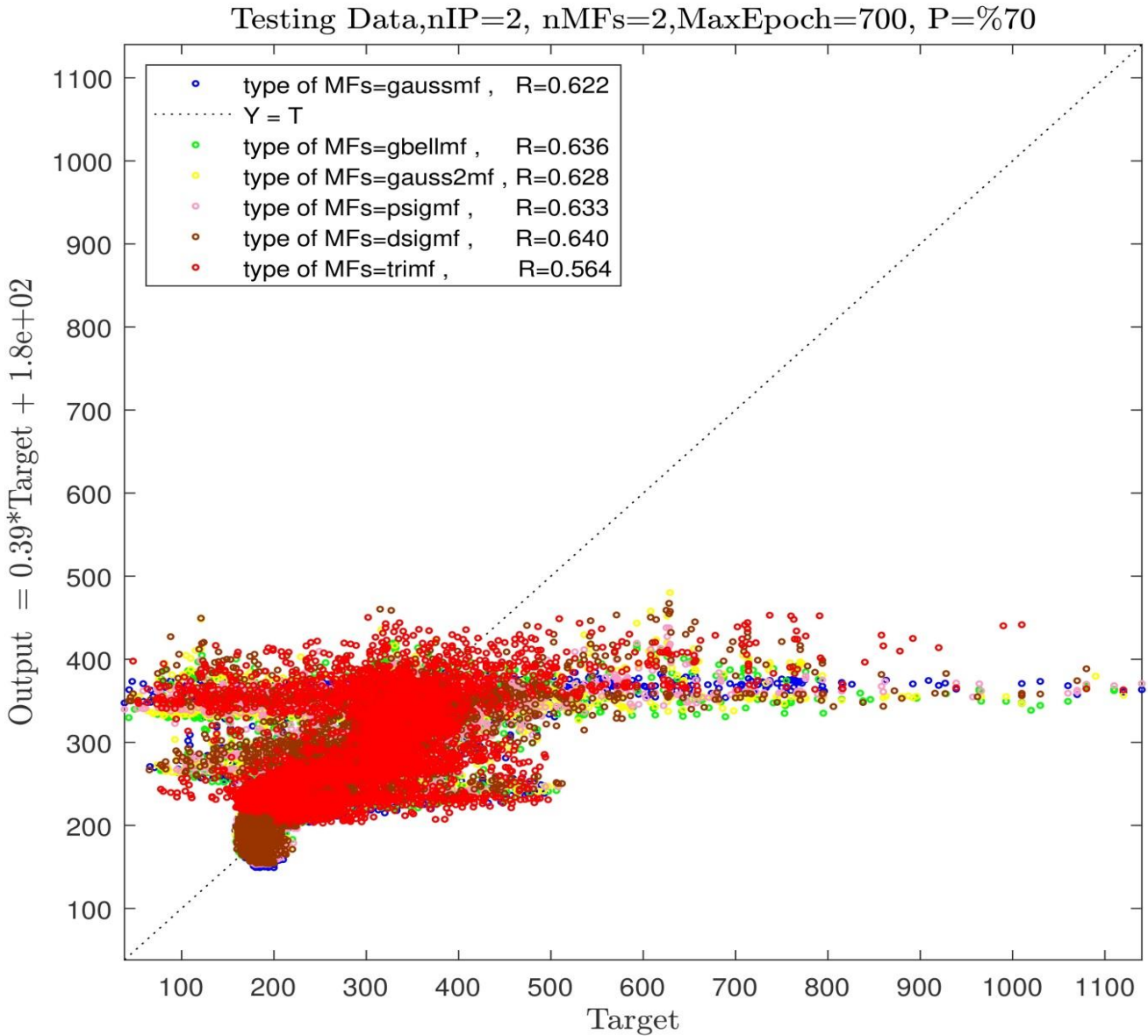

Figure 5(b): Testing process, two inputs, number of MFs=2, various types of MFs.

By performing training and testing processes separately for each type of MFs, according to Figs. 6a and 6b, an increase was observed in the ANFIS intelligence, and $R^2$ rose by 22.6% from 0.64 to 0.86. This growth in ANFIS intelligence sheds light on the positive effect of the increased number of MFs on the improvement of ANFIS intelligence due to a rise in the number of rules. To improve the intelligence as a result of an increase in the number of MFs from four to six, training and testing processes were performed separately. Contrary to the previous section where an increase in the number of MFs from two to four led to a 22% increase in ANFIS intelligence, Figs. 7a and 7b illustrate that increasing the number of MFs from four to six failed to remarkably augment ANFIS intelligence.

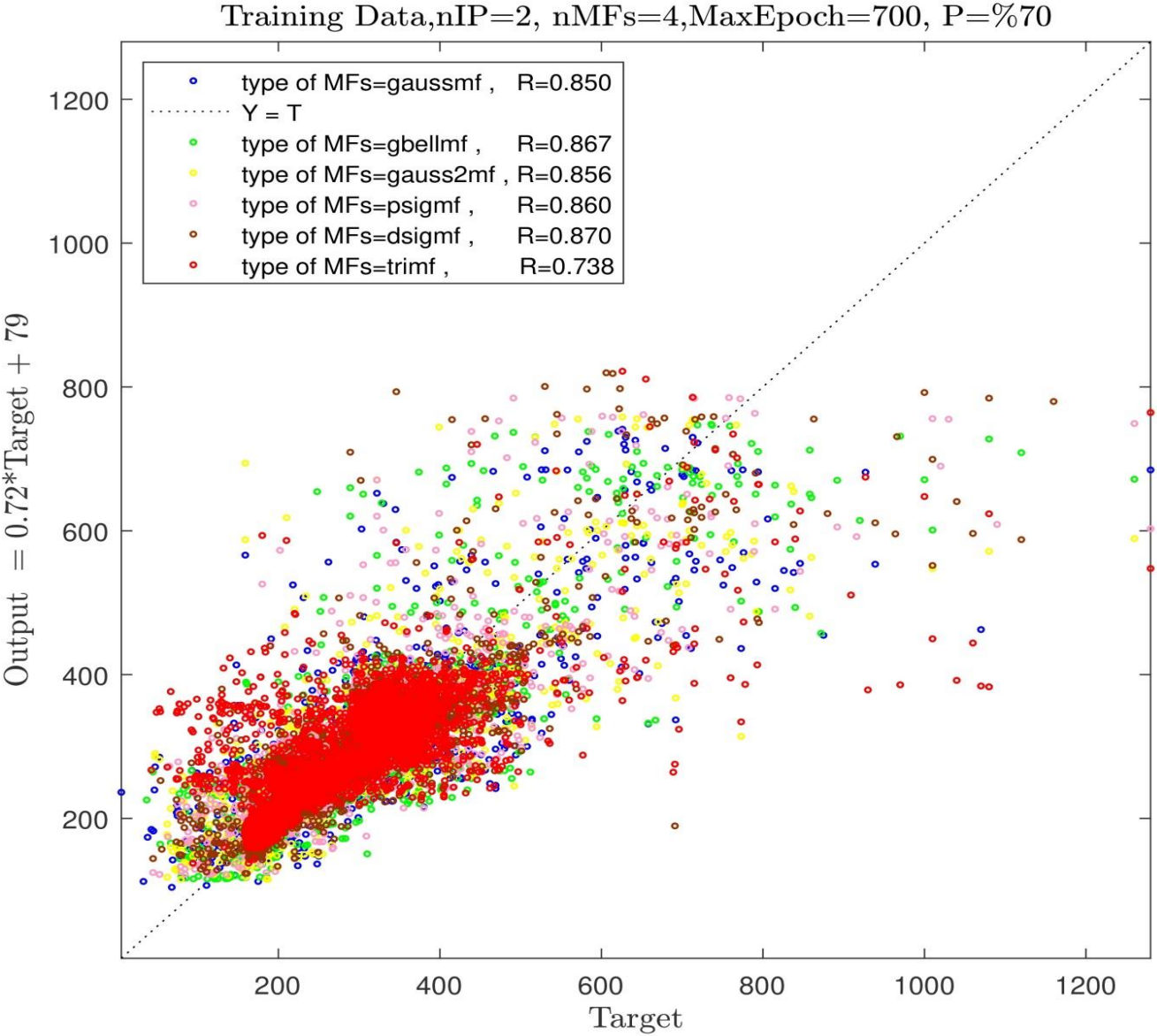

Figure 6(a): Training process, two inputs, number of MFs=4, various types of MFs.

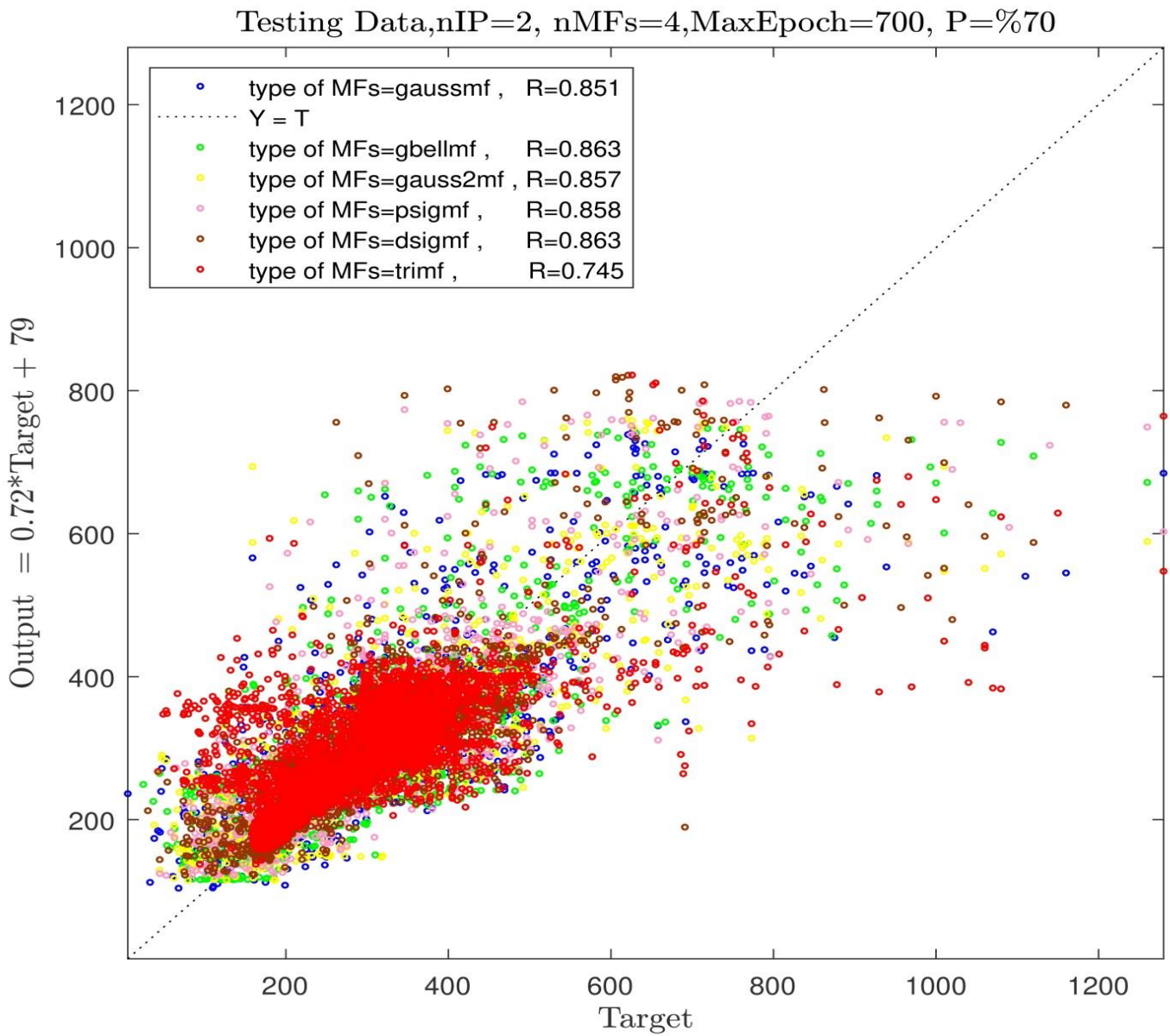

Figure 6(b): Testing process, two inputs, number of MFs=4, various types of MFs.

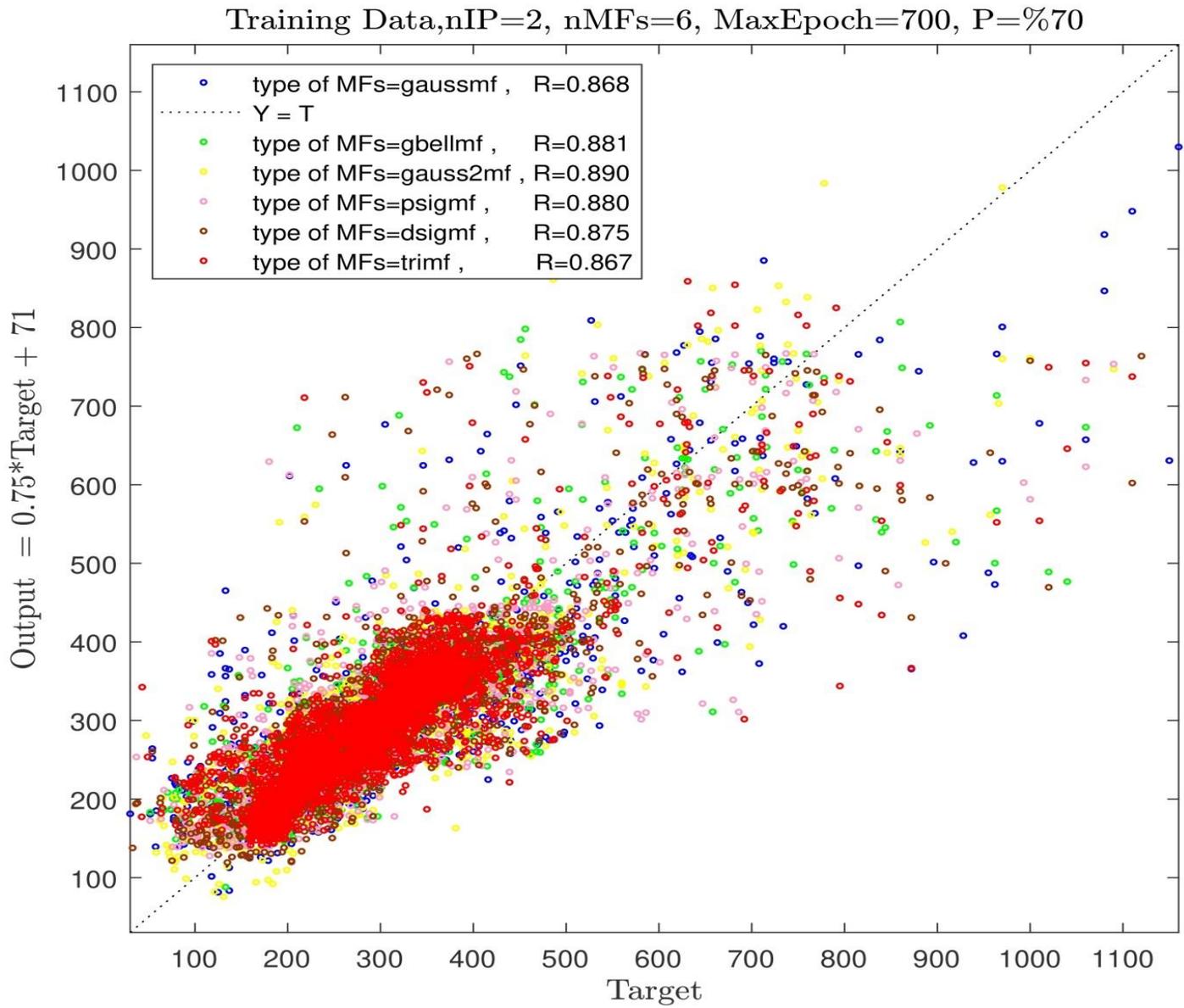

Figure 7(a): Training process, two inputs, number of MFs=6, various types of MFs.

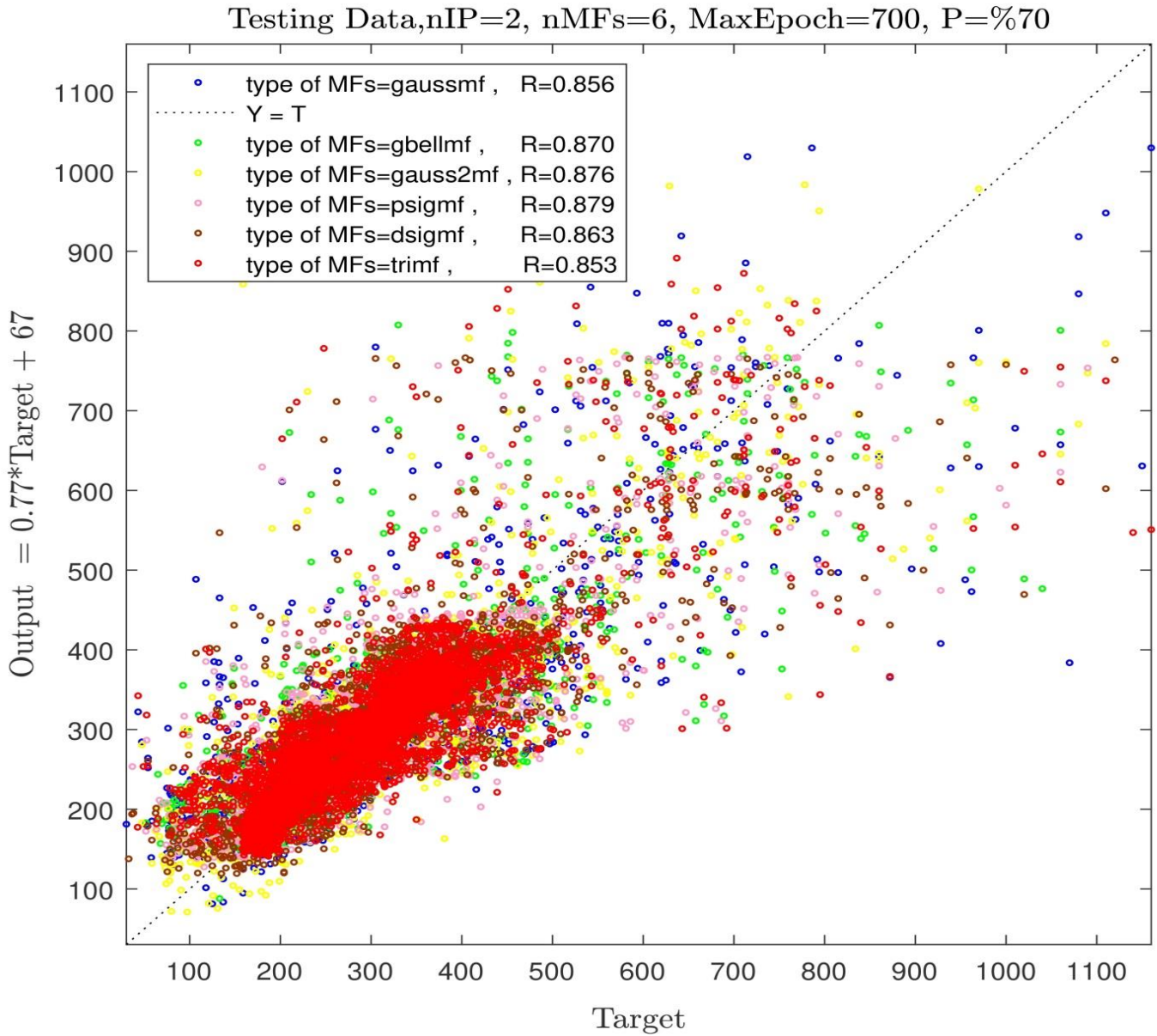

Figure 7(b): Testing process, two inputs, number of MFs=6, various types of MFs.

Different scenarios were evaluated by considering two inputs. Then, the number of inputs was elevated from two to three to promote ANFIS intelligence. To that end, x, y, and z coordinates were considered the inputs, and the pressure gradient was assumed to be the output.

In this condition, training and testing processes were conducted in the case that the number of MFs = 2. The best value obtained for $R^2$ was 0.66, indicating an insignificant improvement compared with the condition that two inputs were used under the same number of MFs (see Figs. 8a and 8b).

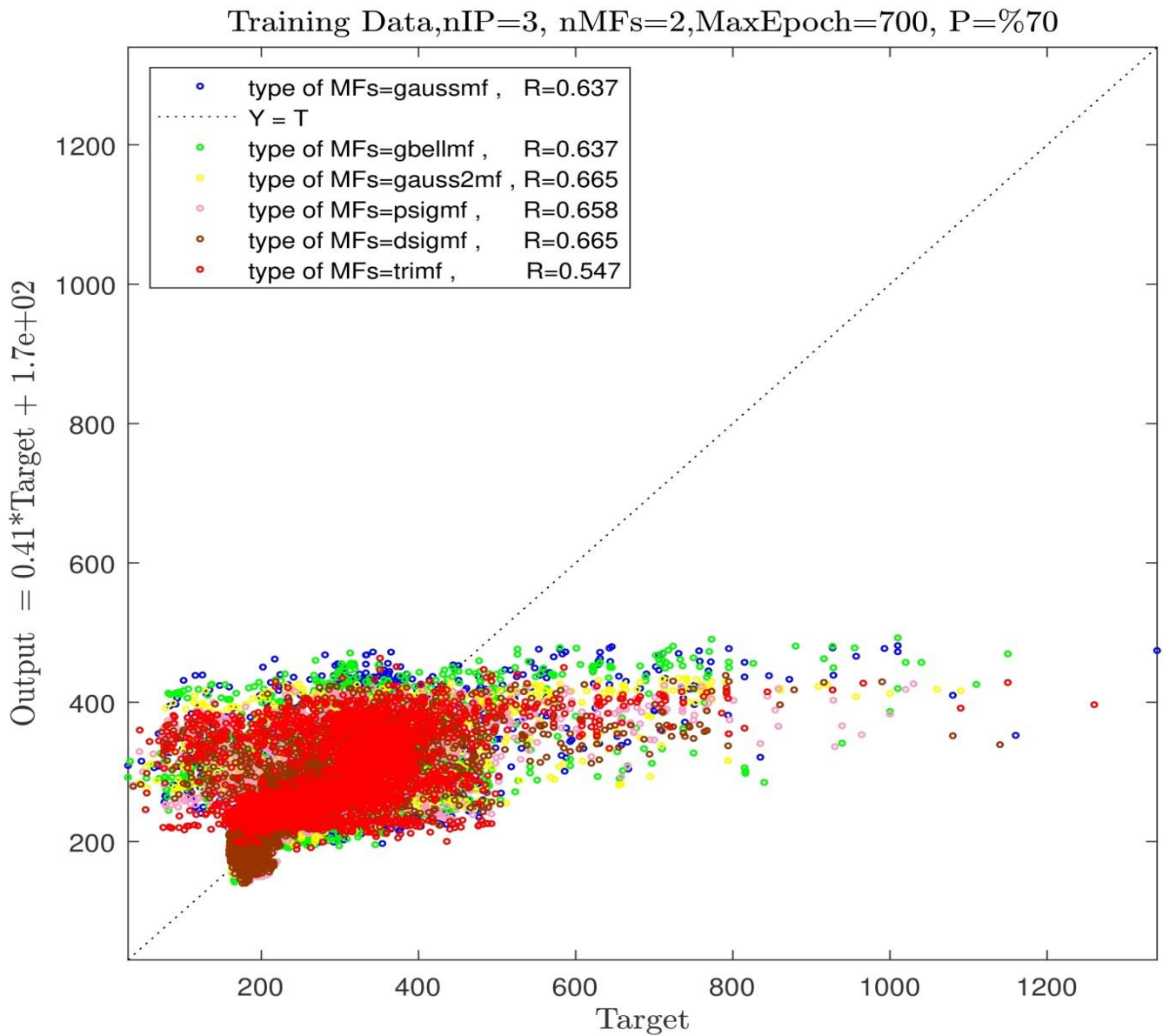

Figure 8(a): Training process, three inputs, number of MFs=2, various types of MFs.

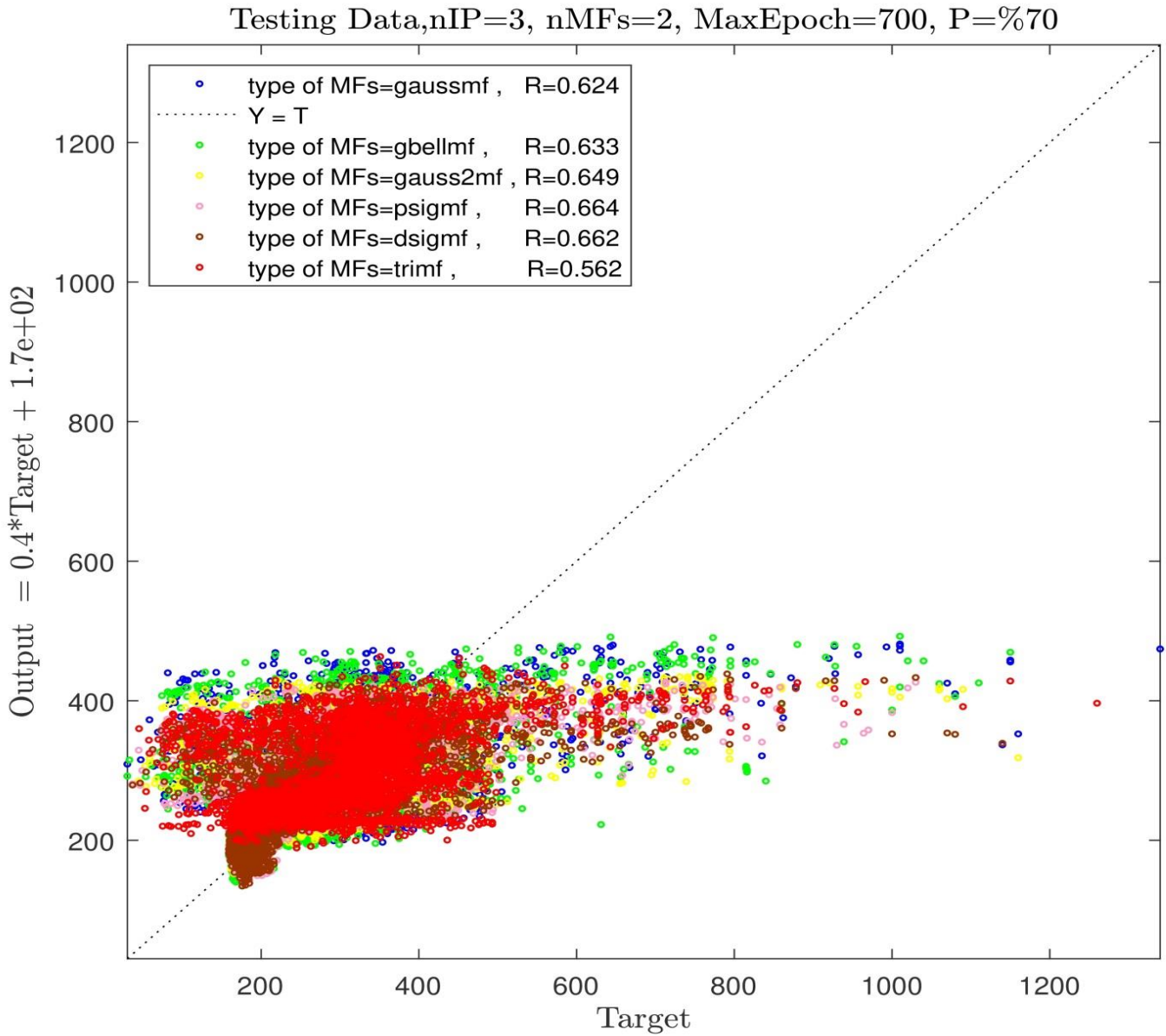

Figure 8(b): Testing process, three inputs, number of MFs=2, various types of MFs.

According to Figs. 9a and 9b, increasing the number of MFs to four, in the case that there are three inputs, leads to extensively satisfying progress in the ANFIS intelligence, and $R^2$ is upped by 28% (0.94), which is a considerably good value. Comparing $R^2$ with the condition where the number of MFs equaled four and that of inputs equaled two, indicates an increase in $R^2$ from 0.86 to 0.94. Such an improvement reflects the positive effect of the increased number of inputs on the promoted level of ANFIS intelligence. Therefore, the number of inputs was decided to increase from three to four. Accordingly, air superficial velocity was added to the system as the fourth input in tandem with other three inputs (viz., x, y, z coordinates), and the pressure gradient was assumed to be the output. Then, the training and testing processes with two MFs were separately accomplished for all types of MFS (see Figs. 10a and 10b).

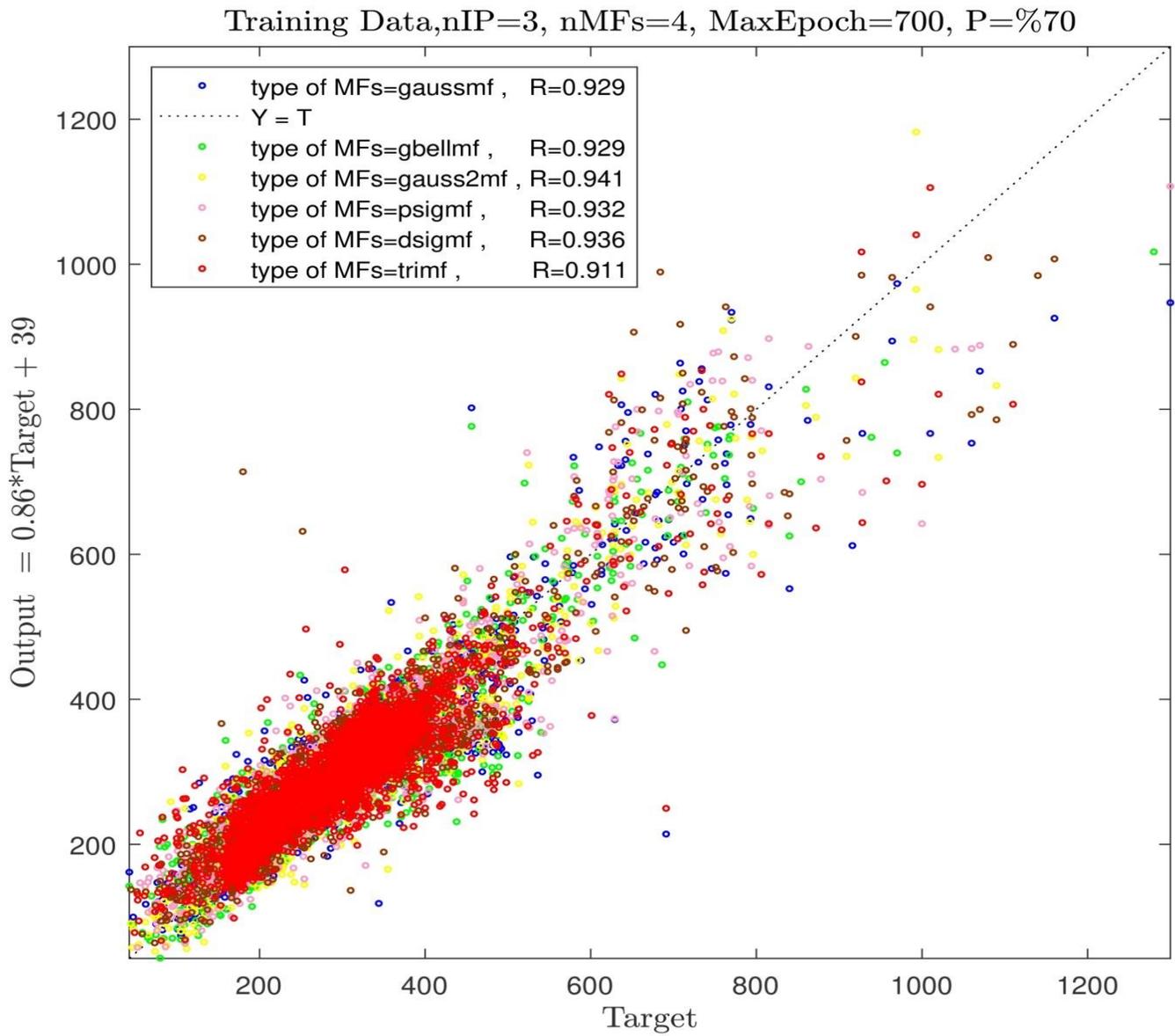

Figure 9(a): Training process, three inputs, number of MFs=4, various types of MFs.

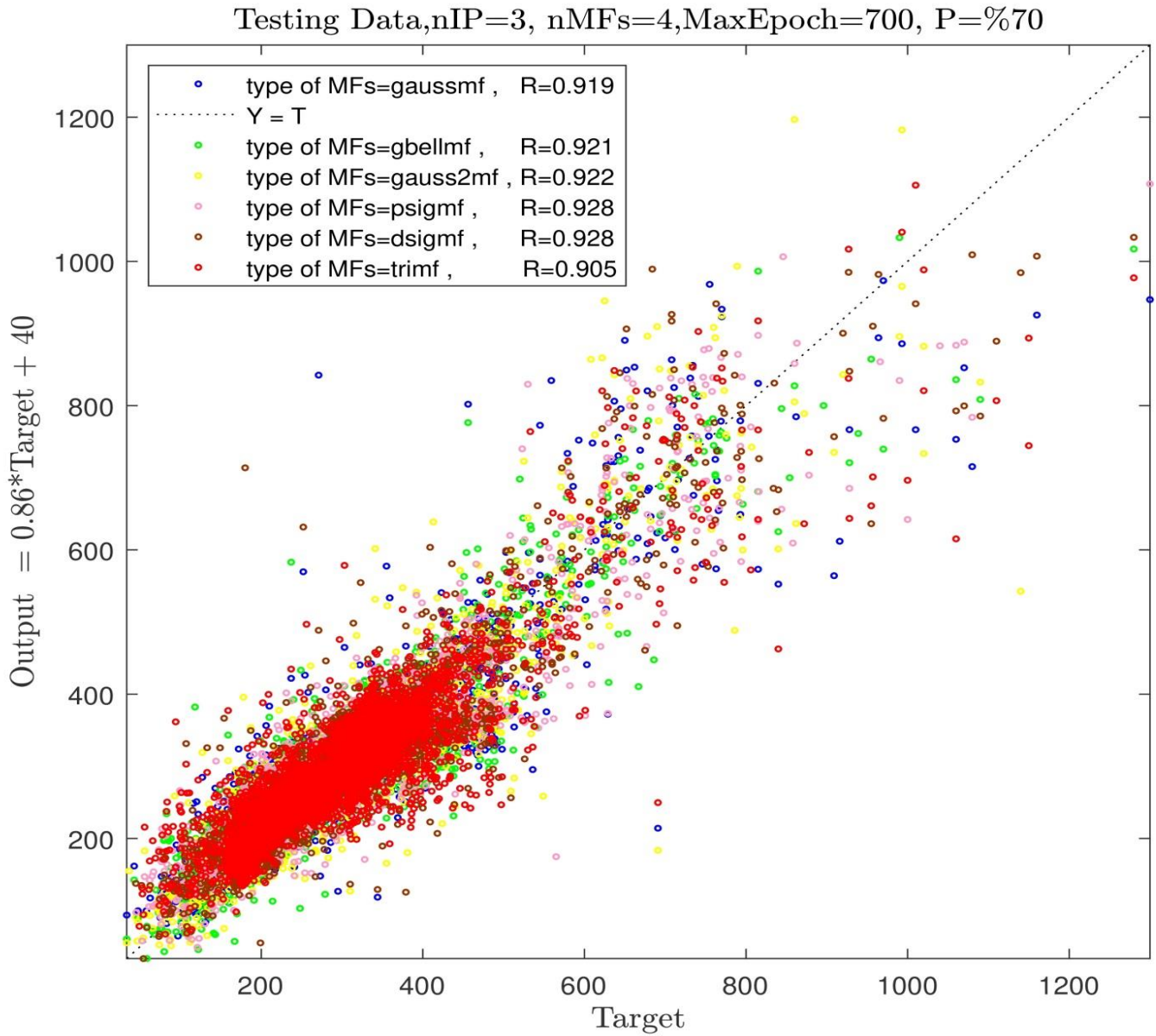

Figure 9(b): Testing process, three inputs, number of MFs=4, various types of MFs.

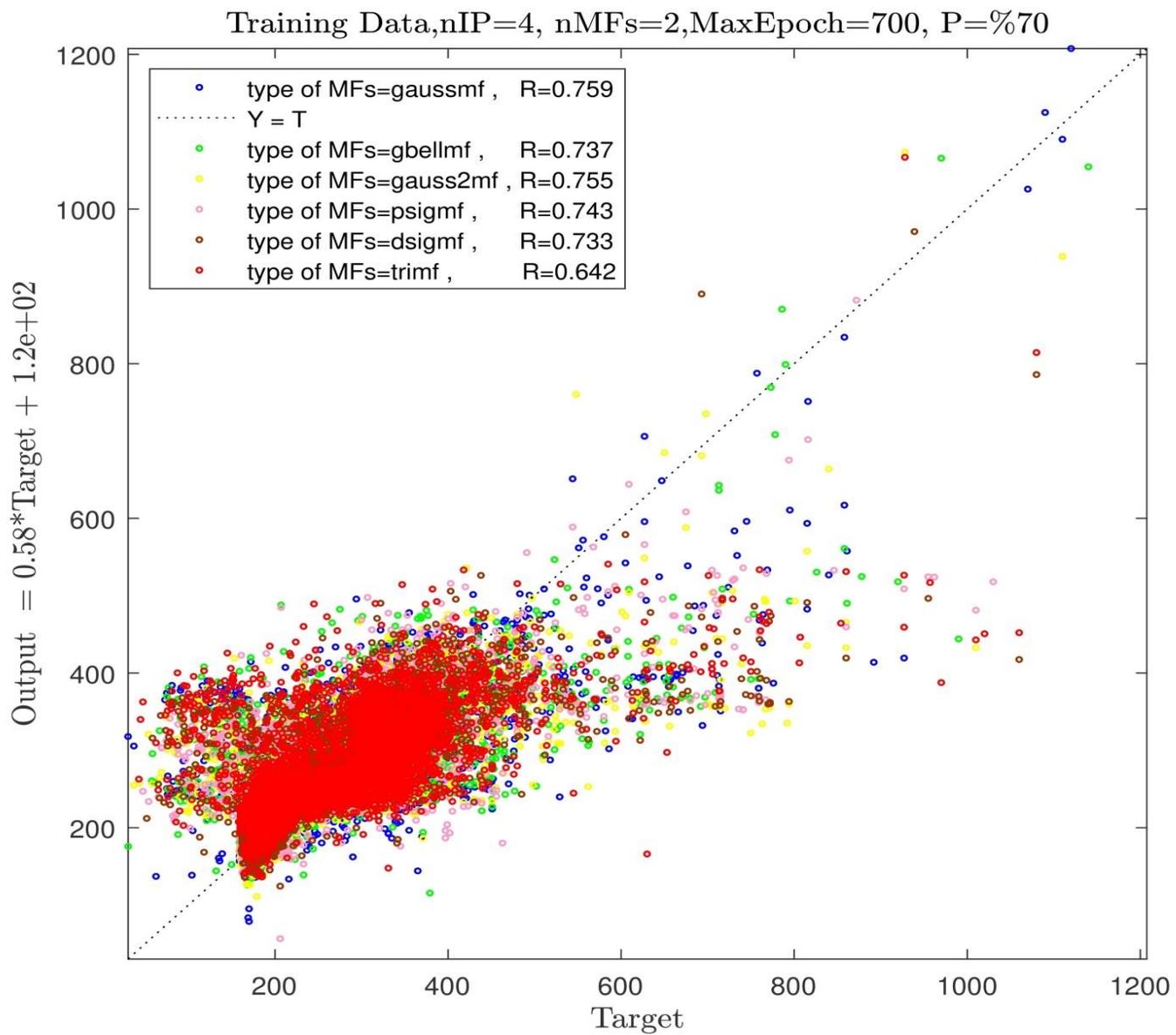

Figure 10(a): Training process, four inputs, number of MFs=2, various types of MFs.

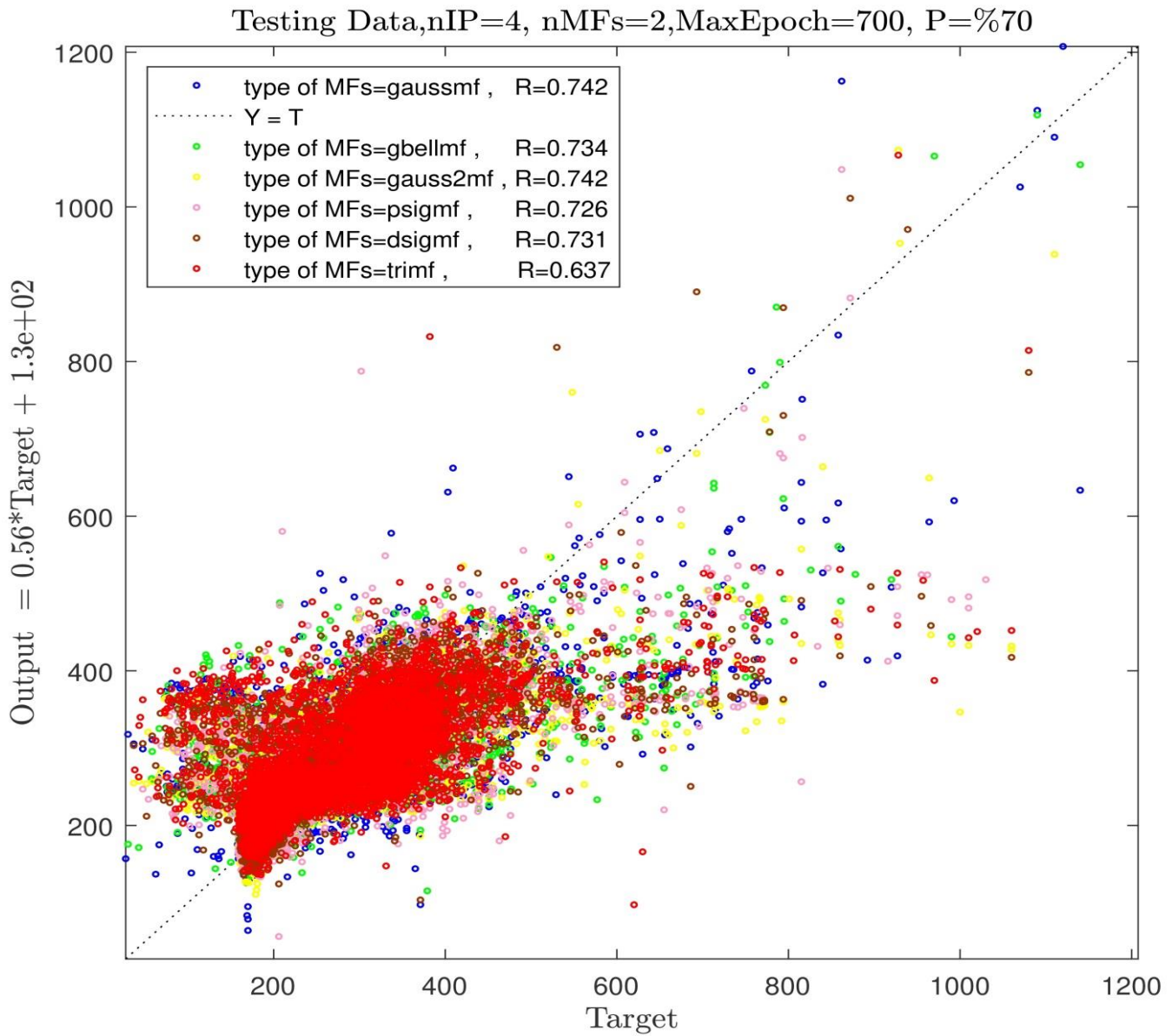

Figure 10(b): Testing process, four inputs, number of MFs=2, various types of MFs.

An improvement in the $R^2$ value (0.76) reflects the noticeably positive effect of the increased number of inputs (from 2 and 3 to 4) on the increased level of system intelligence by 10%. Increasing the number of MFS to four can lead to the complete intelligence of the ANFIS method, such that the $R^2$ value for training and testing processes was obtained to be 0.97 and 0.95, respectively (Figs. 11a and 11b). This

$R^2$ value reflects the considerably high intelligence of the ANFIS method and a significantly appropriate agreement between the outputs of the ANFIS method and that of the CFD method (see Fig. 12 (a, b, c, d, e and f)).

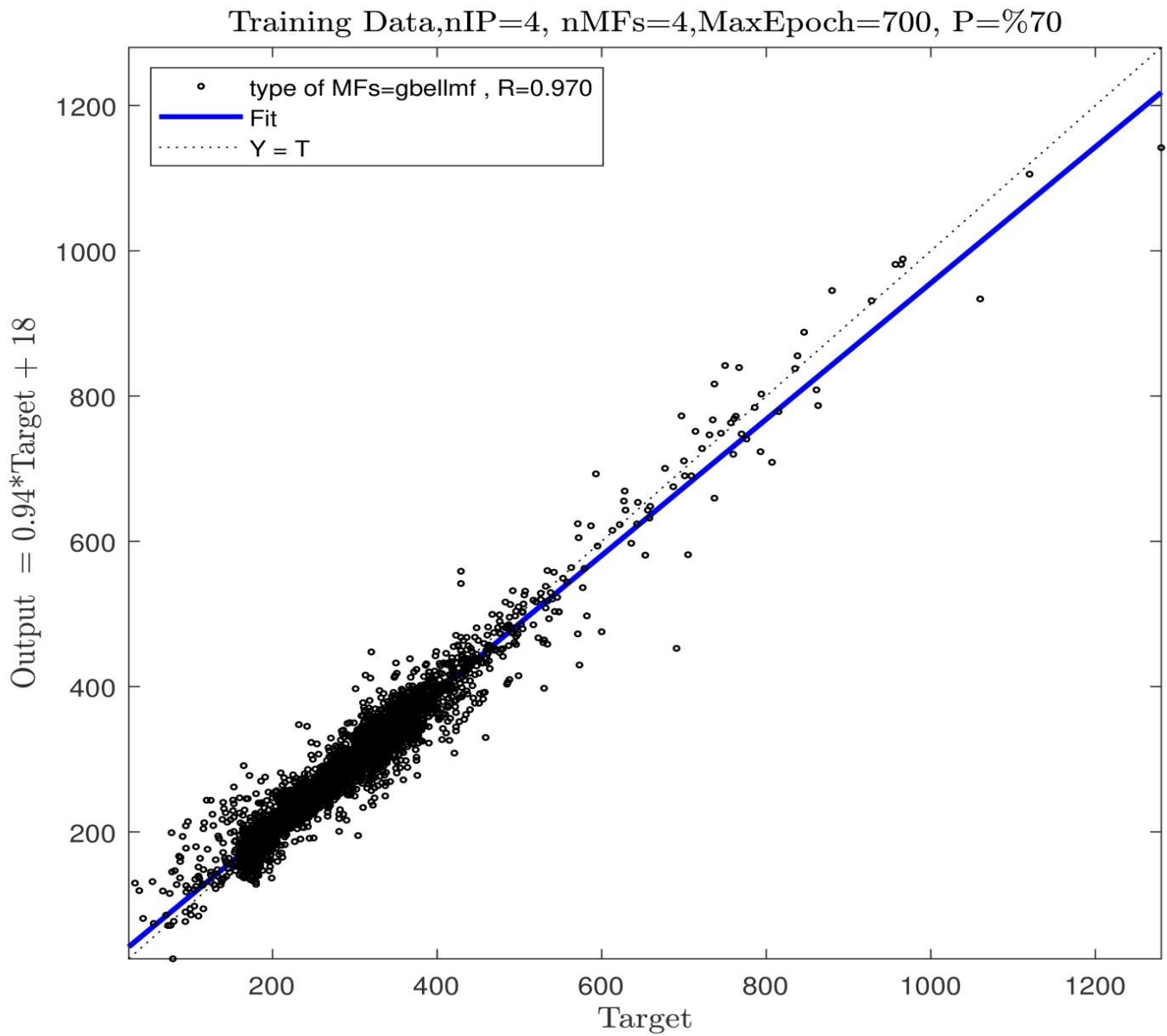

Figure 11(a): Training process, four inputs, number of MFs=4, types of MFs=gbellmf.

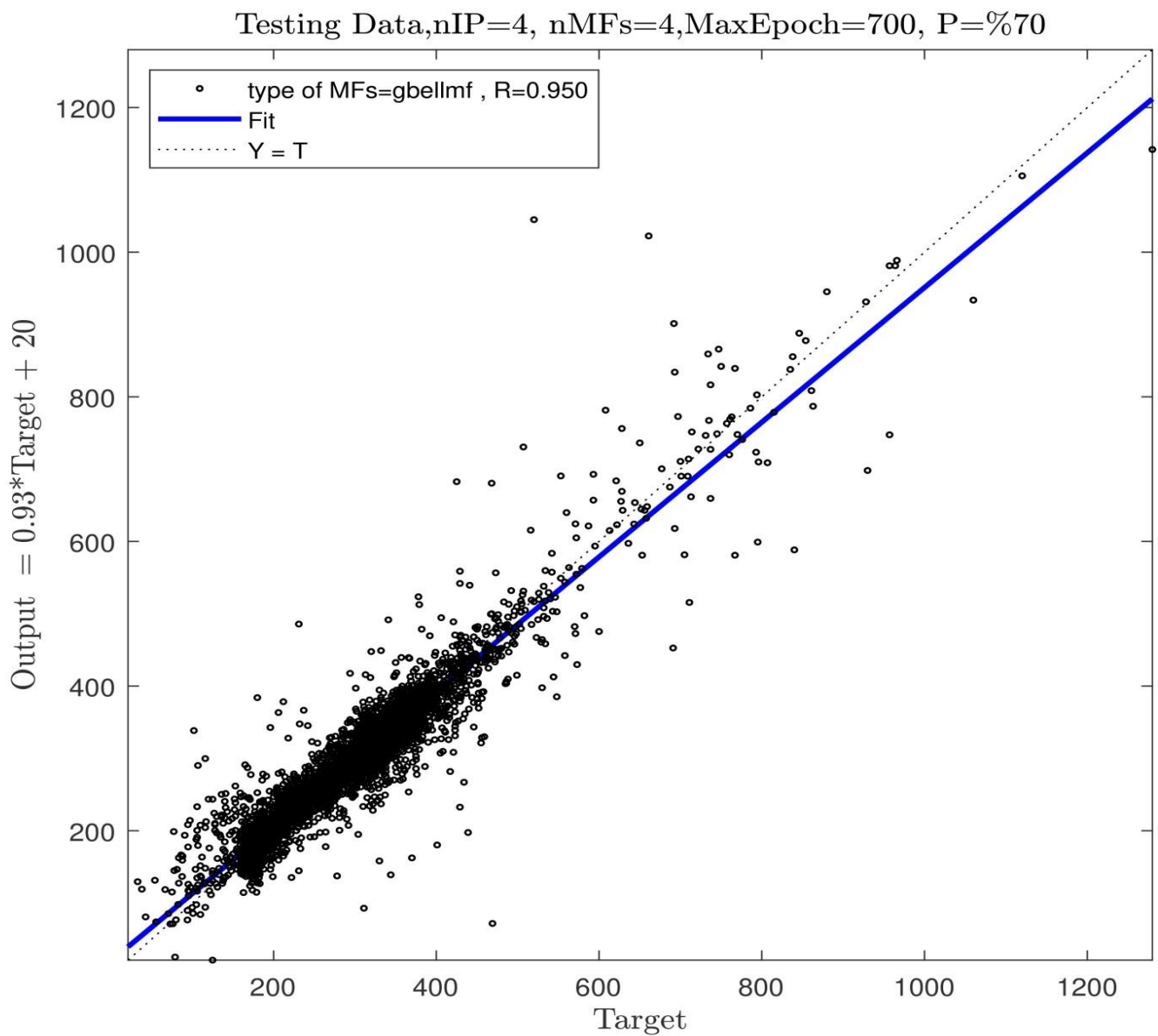

Figure 11(b): Testing process, four inputs, number of MFs=4, types of MFs=gbellmf.

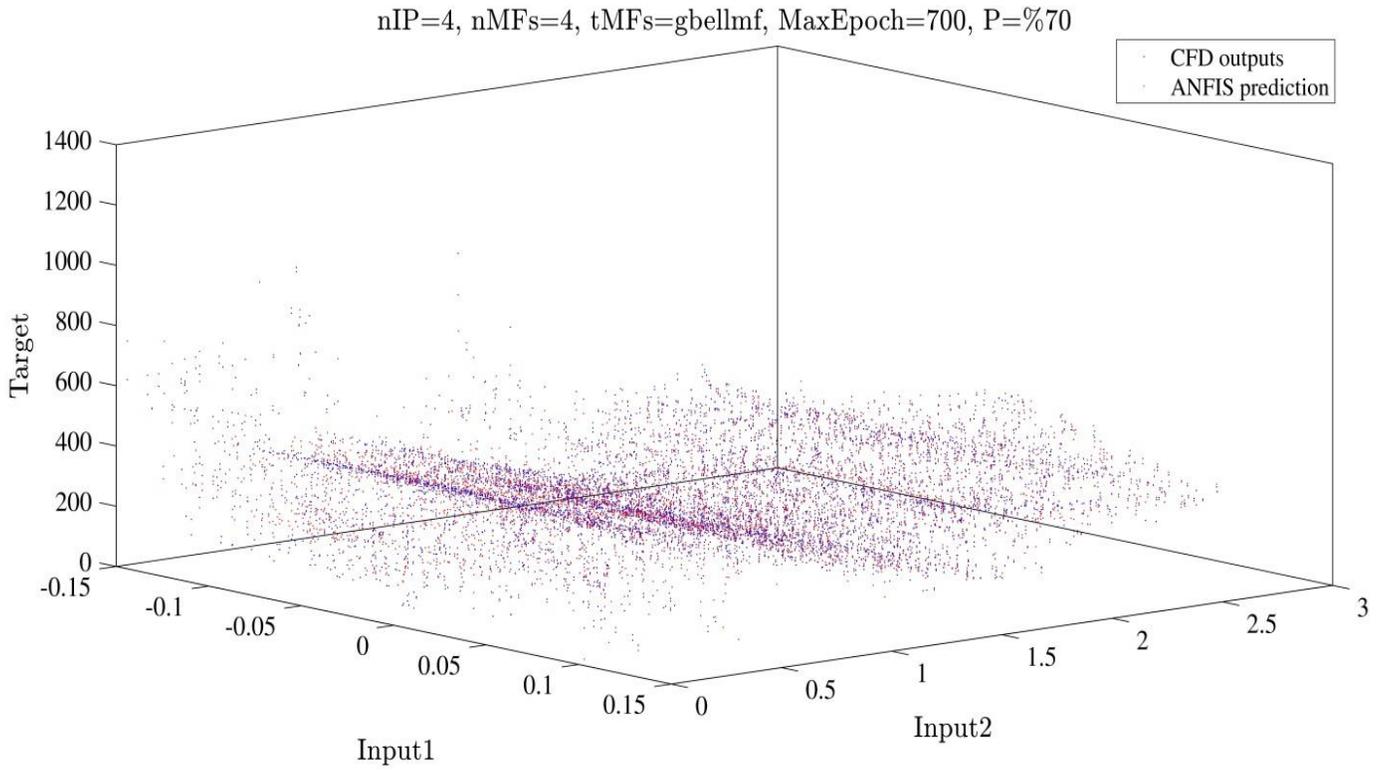

Figure 12(a): Compare ANFIS prediction and CFD output (inputs 1 and 2).

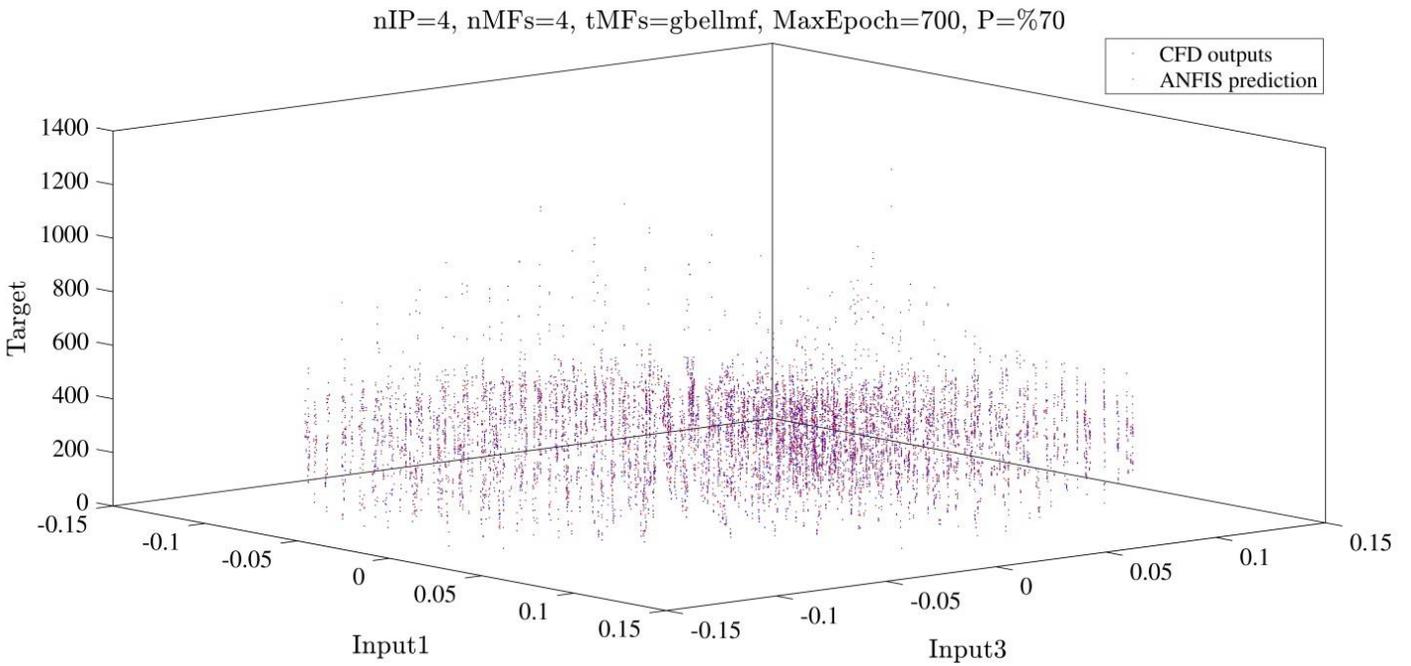

Figure 12(b): Compare ANFIS prediction and CFD output (inputs 1 and 3).

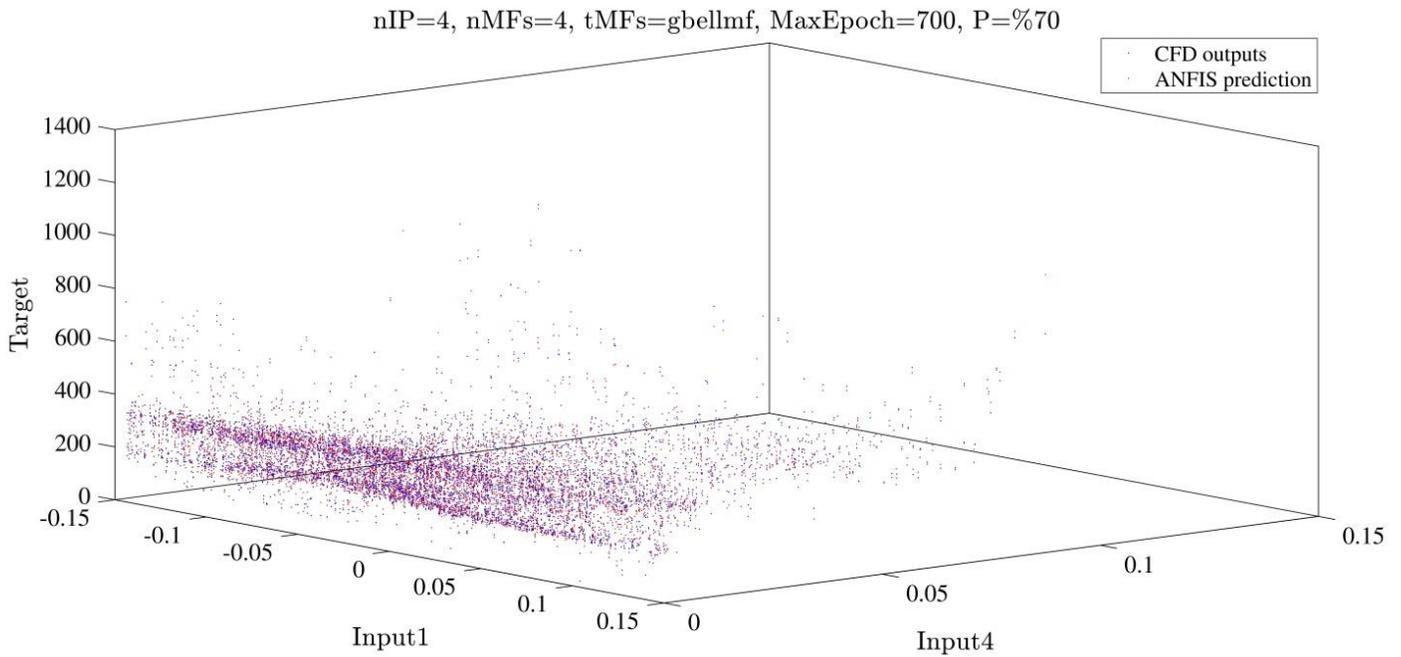

Figure 12(c): Compare ANFIS prediction and CFD output (inputs 1 and 4).

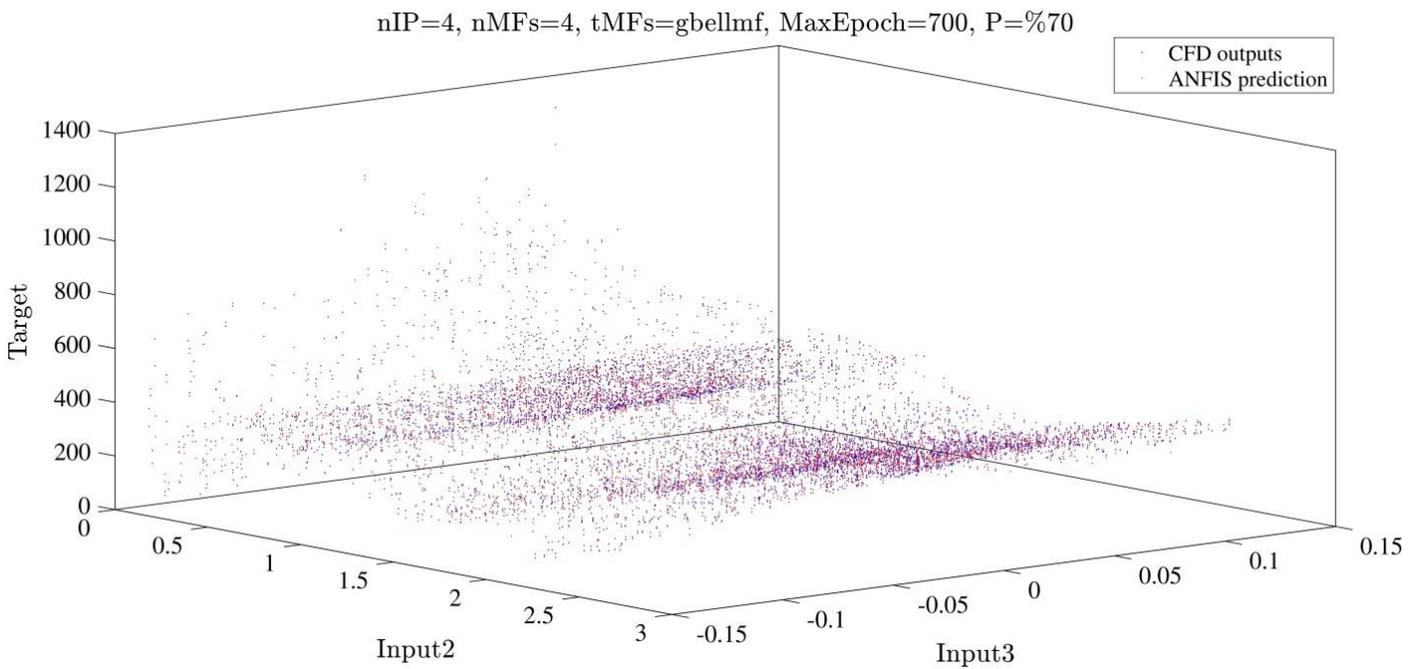

Figure 12(d): Compare ANFIS prediction and CFD output (inputs 2 and 3).

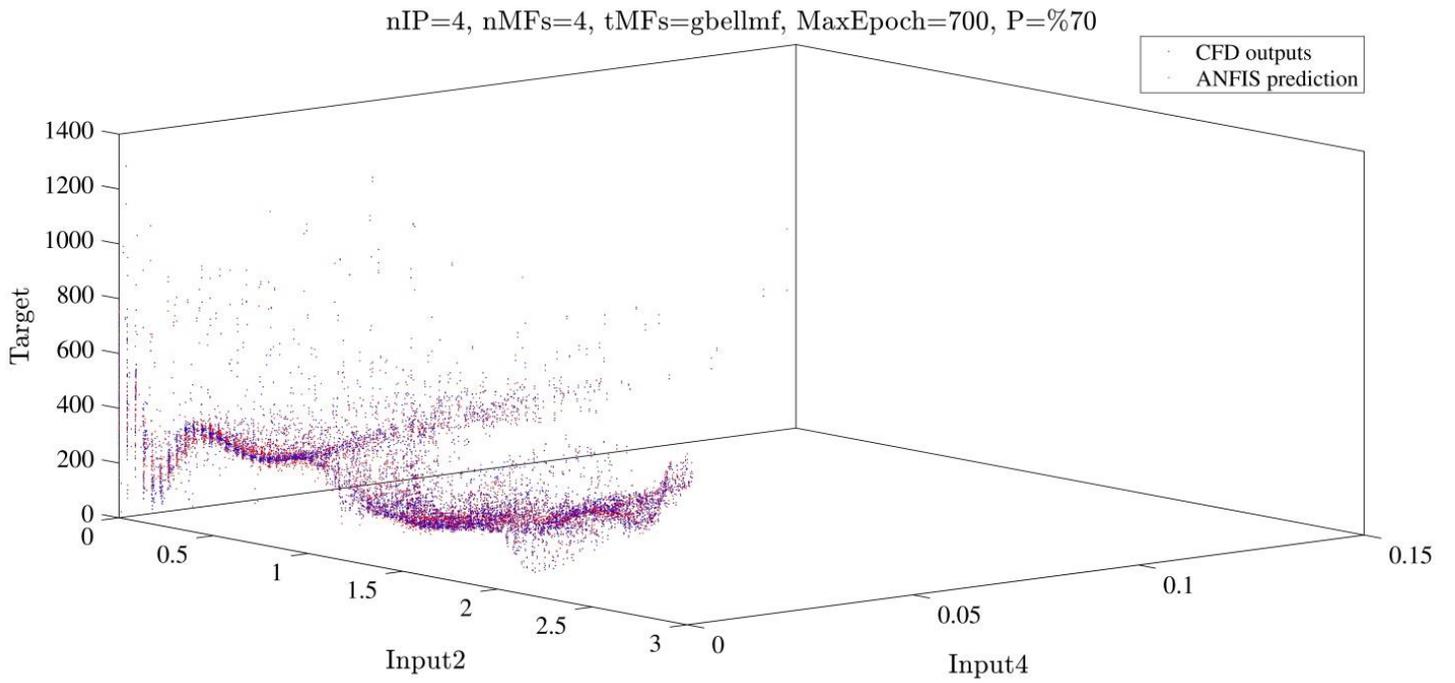

Figure 12(e): Compare ANFIS prediction and CFD output (inputs 2 and 4).

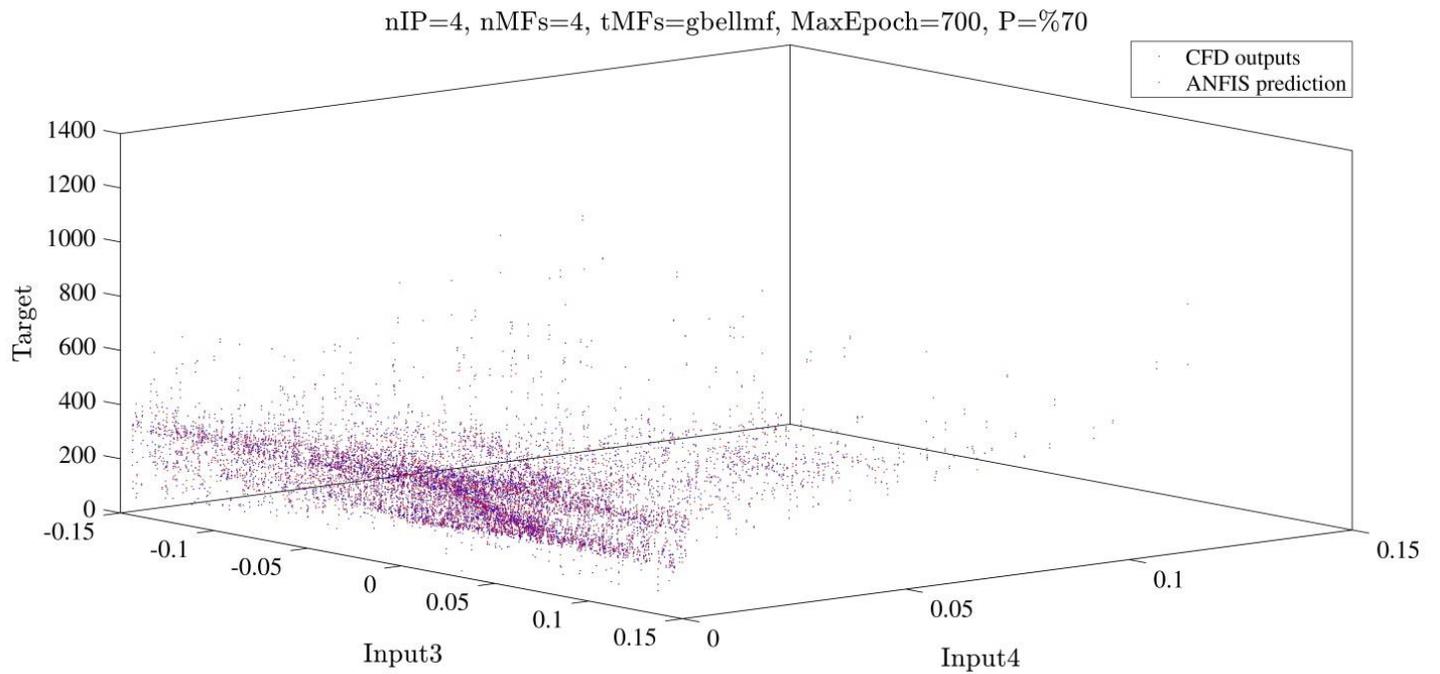

Figure 12(f): Compare ANFIS prediction and CFD output (inputs 3 and 4).

Using the air superficial velocity as the input yielded considerably good results associated with intelligence, enabling different parts of BCR can also be predicted owing to the resulting ANFIS intelligence. Fig. 13 depicts the parts of the BCR that were involved in the learning process .

The ANFIS intelligence can help prognosticate the parts failing to be involved in the learning process, which indicates the significantly high prediction power of the AI (see Fig. 14 (a, b, c, d, and e)).

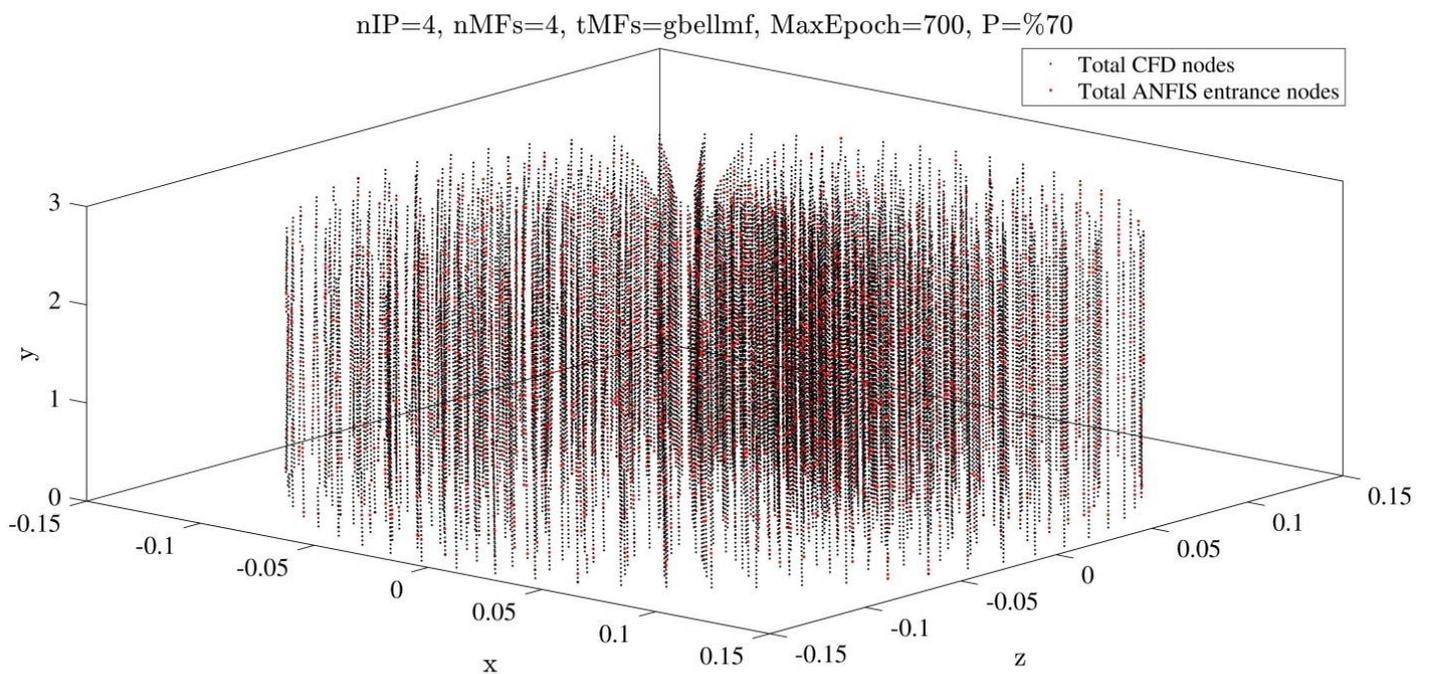

Figure 13: The bubble column nodes present in the ANFIS process.

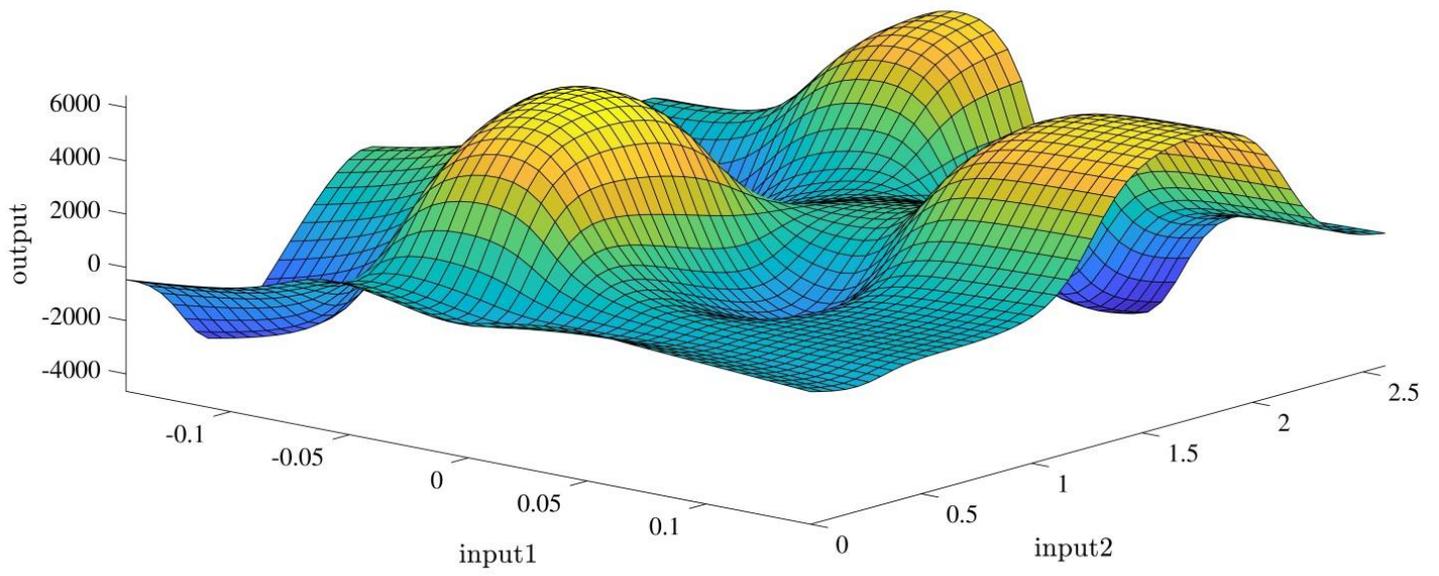

Figure 14(a): ANFIS prediction (inputs 1 and 2).

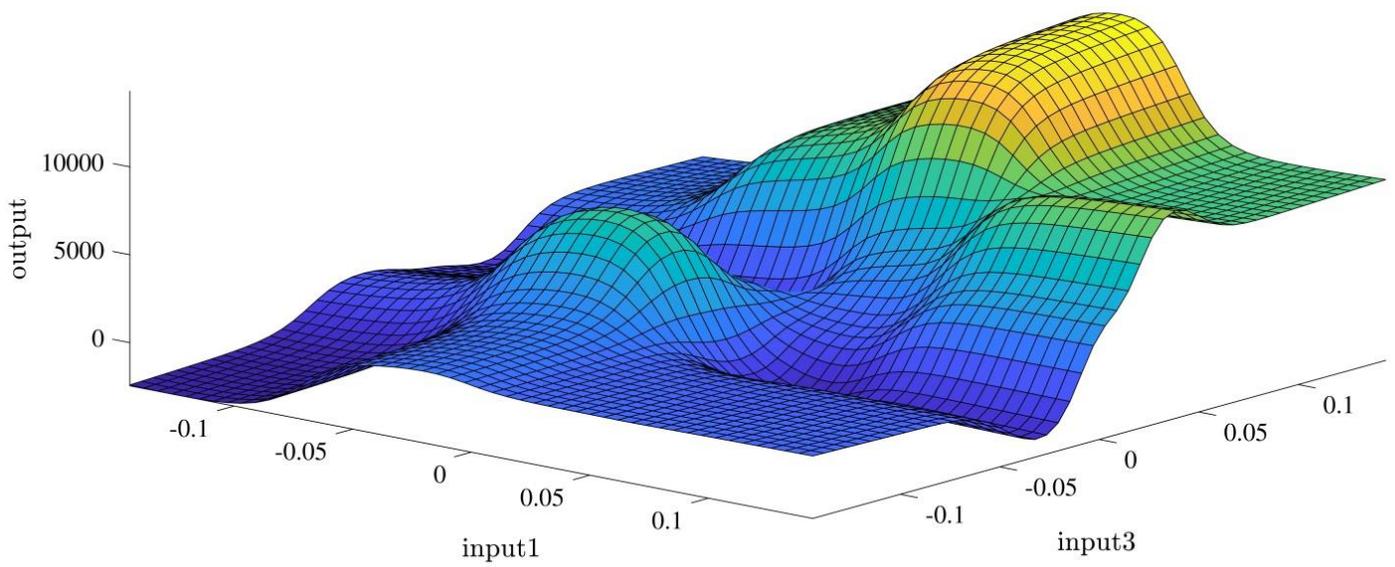

Figure 14(b): ANFIS prediction (inputs 1 and 3).

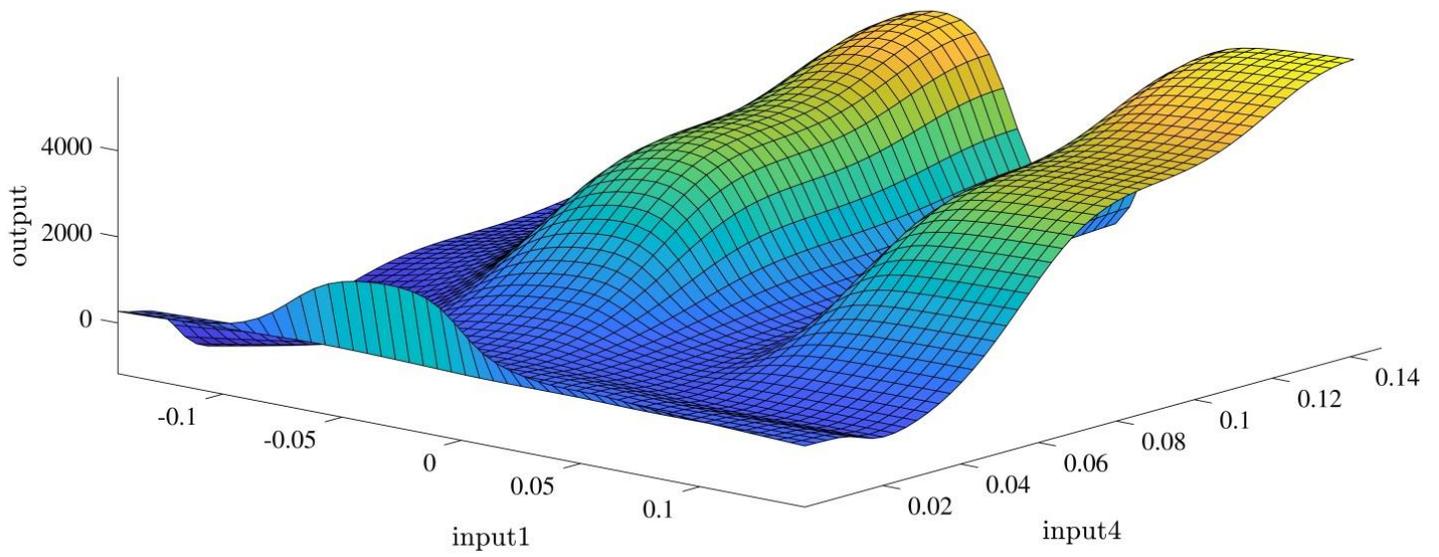

Figure 14(c): ANFIS prediction (inputs 1 and 4).

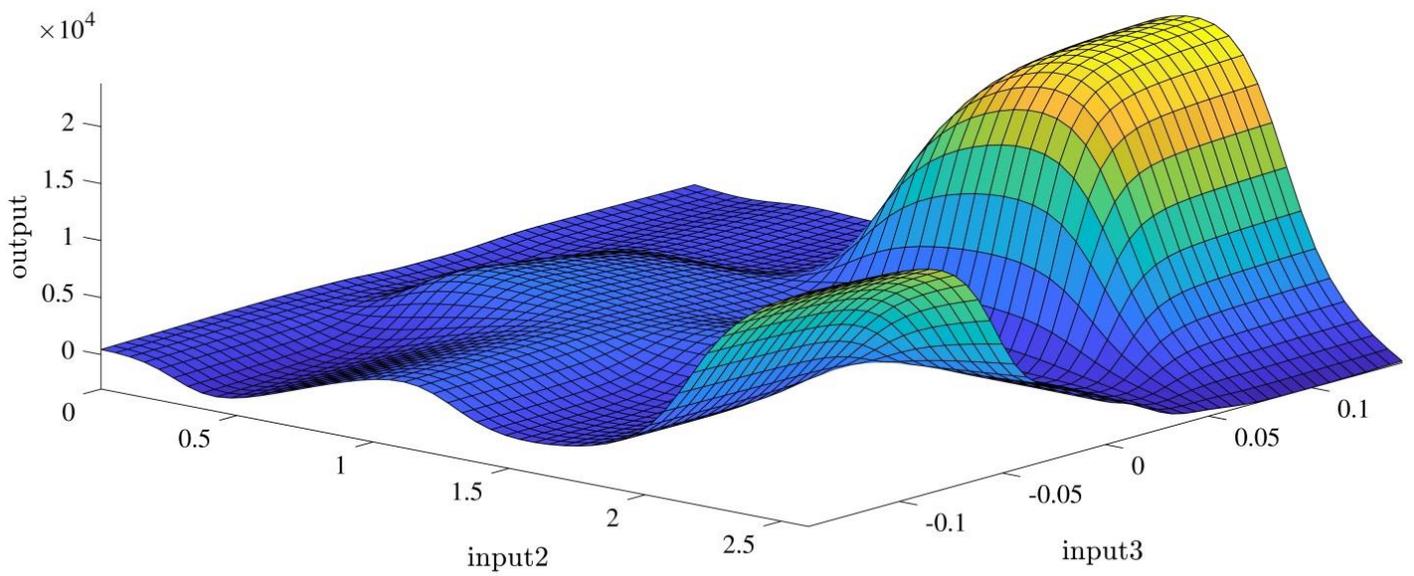

Figure 14(d): ANFIS prediction (inputs 2 and 3).

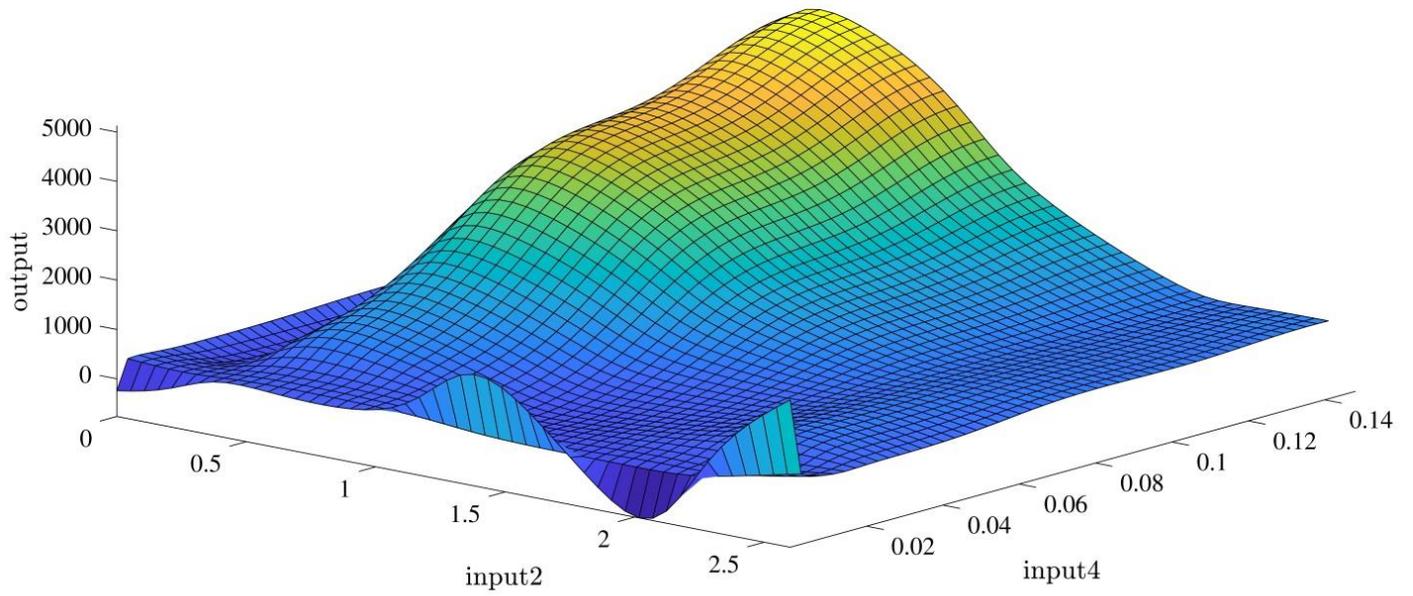

Figure 14(e): ANFIS prediction (inputs 2 and 4).

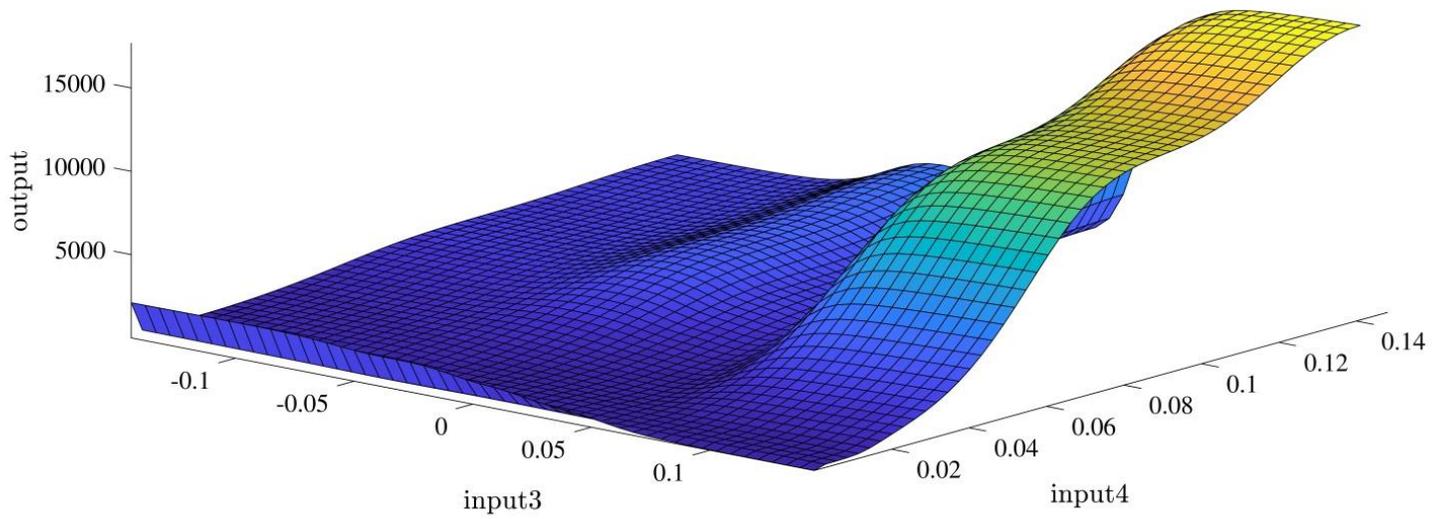

Figure 14(f): ANFIS prediction (inputs 3 and 4).

Integrating AI (ANFIS method) with the CFD method reduces the computational time required by the CFD method and also avoids solving complex equations by the CFD. Taking advantage of the ANFIS intelligence can provide further information about considerably more points.

## 4. Conclusions

This study uses the new framework of the ANFIS plus CFD method to predict the pressure gradient as an output of the ANFIS with combined input and output hydrodynamics parameters of the cylindrical bubble column reactor (such as mesh positions and gas speed) as the ANFIS input parameters. For this study, different function characteristics, the number of function and input parameters are used to achieve the accurate ANFIS method for the prediction of reactor. The results show that the input parameters and the number of rules significantly affect the accuracy of the intelligent algorithm. Due to the high density of neural nodes for high input parameters or the high number of rules the algorithms can mimic the pressure gradient in the bubble column reactor. The membership function has a minimal effect on the intelligent of the combination of the ANFIS and CFD. Furthermore, we suggest that to improve the accuracy of the method after using the maximum input parameters and number of rules, the tuning parameters should be considered. Additionally, the results show that the output parameters such as the speed of gas in the column can be used during the learning process, and for the better understanding of this complex hydrodynamics behavior in the column, deep learning methods such as Long short-term memory (LSTM) (Vargas et al. 2017) can be considered.

# References

Abdulshahed, A. M., Longstaff, A. P., & Fletcher, S. (2015). The application of ANFIS prediction models for thermal error compensation on CNC machine tools. *Applied Soft Computing, 27*, 158-168.

Ali Ghorbani, M., Kazempour, R., Chau, K.-W., Shamshirband, S., & Taherei Ghazvinei, P. (2018). Forecasting pan evaporation with an integrated artificial neural network quantum-behaved particle swarm optimization model: A case study in Talesh, Northern Iran. *Engineering Applications of Computational Fluid Mechanics, 12*(1), 724-737.

Anabtawi, M., Abu-Eishah, S., Hilal, N., & Nabhan, M. (2003). Hydrodynamic studies in both bi-dimensional and three-dimensional bubble columns with a single sparger. *Chemical Engineering and Processing: Process Intensification, 42*(5), 403-408.

Avila, G., & Pacheco-Vega, A. (2009). Fuzzy-C-Means-Based Classification of Thermodynamic-Property Data: A Critical Assessment. *Numerical Heat Transfer, Part A: Applications, 56*(11), 880-896.

Azwadi, C., Razzaghian, M., Pourtousi, M., & Safdari, A. (2013). Numerical prediction of free convection in an open ended enclosure using lattice Boltzmann numerical method. *Int. J. Mech. Mater. Eng, 8*, 58-62.

Azwadi, C. S. N., Zeinali, M., Safdari, A., & Kazemi, A. (2013). Adaptive-network-based fuzzy inference system analysis to predict the temperature and flow fields in a lid-driven cavity. *Numerical Heat Transfer, Part A: Applications, 63*(12), 906-920.

Babanezhad, M., Rezakazemi, M., Hajilary, N., & Shirazian, S. Liquid-phase chemical reactors: Development of 3D hybrid model based on CFD-adaptive network-based fuzzy inference system. *The Canadian Journal of Chemical Engineering*.

Behkish, A., Men, Z., Inga, J. R., & Morsi, B. I. (2002). Mass transfer characteristics in a large-scale slurry bubble column reactor with organic liquid mixtures. *Chemical Engineering Science, 57*(16), 3307-3324.

Besagni, G., Guédon, G. R., & Inzoli, F. (2016). Annular Gap Bubble Column: Experimental Investigation and Computational Fluid Dynamics Modeling. *Journal of Fluids Engineering, 138*(1), 011302.

Besbes, S., El Hajem, M., Aissia, H. B., Champagne, J., & Jay, J. (2015). PIV measurements and Eulerian–Lagrangian simulations of the unsteady gas–liquid flow in a needle sparger rectangular bubble column. *Chemical Engineering Science, 126*, 560-572.

Bouaifi, M., Hebrard, G., Bastoul, D., & Roustan, M. (2001). A comparative study of gas hold-up, bubble size, interfacial area and mass transfer coefficients in stirred gas–liquid reactors and bubble columns. *Chemical Engineering and Processing: Process Intensification, 40*(2), 97-111.

Burns, A. D., Frank, T., Hamill, I., & Shi, J.-M. (2004). *The Favre averaged drag model for turbulent dispersion in Eulerian multi-phase flows.* Paper presented at the 5th international conference on multiphase flow, ICMF.

Buwa, V. V., Deo, D. S., & Ranade, V. V. (2006). Eulerian–Lagrangian simulations of unsteady gas–liquid flows in bubble columns. *International Journal of Multiphase Flow, 32*(7), 864-885.

Buwa, V. V., & Ranade, V. V. (2002). Dynamics of gas–liquid flow in a rectangular bubble column: experiments and single/multi-group CFD simulations. *Chemical Engineering Science, 57*(22-23), 4715-4736.

Buwa, V. V., & Ranade, V. V. (2003). Mixing in bubble column reactors: role of unsteady flow structures. *The Canadian Journal of Chemical Engineering, 81*(3-4), 402-411.

Chau, K.-w. (2017). Use of meta-heuristic techniques in rainfall-runoff modelling: Multidisciplinary Digital Publishing Institute.

Chen, W., Hasegawa, T., Tsutsumi, A., Otawara, K., & Shigaki, Y. (2003). Generalized dynamic modeling of local heat transfer in bubble columns. *Chemical Engineering Journal, 96*(1-3), 37-44.

Cho, Y. J., Woo, K. J., Kang, Y., & Kim, S. D. (2002). Dynamic characteristics of heat transfer coefficient in pressurized bubble columns with viscous liquid medium. *Chemical Engineering and Processing: Process Intensification, 41*(8), 699-706.

Choubin, B., Borji, M., Mosavi, A., Sajedi-Hosseini, F., Singh, V.P. and Shamshirband, S., 2019. Snow avalanche hazard prediction using machine learning methods. Journal of Hydrology, p.123929.

Choubin, B., Moradi, E., Golshan, M., Adamowski, J., Sajedi-Hosseini, F. and Mosavi, A., 2019. An Ensemble prediction of flood susceptibility using multivariate discriminant analysis, classification and regression trees, and support vector machines. Science of the Total Environment, 651, pp.2087-2096.

Chuntian, C., & Chau, K.-W. (2002). Three-person multi-objective conflict decision in reservoir flood control. *European Journal of Operational Research, 142*(3), 625-631.

Clift, R. (1978). Bubbles. *Drops and Particles*.

de Bertodana, M. L. (1992). *Turbulent bubbly two-phase flows in a triangular.* PhD Thesis, Renssealaer Polytechnic Institute, Troy.

de Bertodano, M. A. L. (1992). *Turbulent bubbly two-phase flow in a triangular duct.*

Deen, N. G., Solberg, T., & Hjertager, B. H. (2000). *Numerical simulation of the gas-liquid flow in a square cross-sectioned bubble column.* Paper presented at the Proceedings of 14th Int. Congress of Chemical and Process Engineering: CHISA (Praha, Czech Republic, 2000).

Degaleesan, S., Dudukovic, M., & Pan, Y. (2001). Experimental study of gas-induced liquid-flow structures in bubble columns. *AIChE Journal, 47*(9), 1913-1931.

Dehghani, M., Riahi-Madvar, H., Hooshyaripor, F., Mosavi, A., Shamshirband, S., Zavadskas, E.K. and Chau, K.W., 2019. Prediction of hydropower generation using grey wolf optimization adaptive neuro-fuzzy inference system. *Energies*, *12*(2), p.289.

Dhotre, M., Ekambara, K., & Joshi, J. (2004). CFD simulation of sparger design and height to diameter ratio on gas hold-up profiles in bubble column reactors. *Experimental thermal and fluid science, 28*(5), 407-421.

Dhotre, M. T., Niceno, B., Smith, B. L., & Simiano, M. (2009). Large-eddy simulation (LES) of the large scale bubble plume. *Chemical Engineering Science, 64*(11), 2692-2704.

Díaz, M. E., Iranzo, A., Cuadra, D., Barbero, R., Montes, F. J., & Galán, M. A. (2008). Numerical simulation of the gas–liquid flow in a laboratory scale bubble column: influence of bubble size distribution and non-drag forces. *Chemical Engineering Journal, 139*(2), 363-379.

Ekambara, K., Dhotre, M. T., & Joshi, J. B. (2005). CFD simulations of bubble column reactors: 1D, 2D and 3D approach. *Chemical Engineering Science, 60*(23), 6733-6746.

Essadki, H., Nikov, I., & Delmas, H. (1997). Electrochemical probe for bubble size prediction in a bubble column. *Experimental thermal and fluid science, 14*(3), 243-250.

Fan, L. (1989). Gas-Solid-Liquid Fluidization Engineering: Butterworth, Boston, USA.

Forret, A., Schweitzer, J., Gauthier, T., Krishna, R., & Schweich, D. (2003). Influence of scale on the hydrodynamics of bubble column reactors: an experimental study in columns of 0.1, 0.4 and 1 m diameters. *Chemical Engineering Science, 58*(3-6), 719-724.

Fritzke, B. (1997). *Incremental neuro-fuzzy systems.* Paper presented at the Applications of Soft Computing.

Hills, J. J. T. I. C. E. (1974). Radial non-uniformity of velocity and voidage in a bubble column. *52*, 1-9.

Hyndman, C. L., Larachi, F., & Guy, C. J. C. e. s. (1997). Understanding gas-phase hydrodynamics in bubble columns: a convective model based on kinetic theory. *52*(1), 63-77.

Islam, M. T., Ganesan, P., & Cheng, J. (2015). A pair of bubbles' rising dynamics in a xanthan gum solution: a CFD study. *RSC Advances, 5*(11), 7819-7831.

Jang, J.-S. (1993). ANFIS: adaptive-network-based fuzzy inference system. *IEEE transactions on systems, man, and cybernetics, 23*(3), 665-685.

Jang, J.-S. R., Sun, C.-T., & Mizutani, E. (1997). Neuro-fuzzy and soft computing; a computational approach to learning and machine intelligence.

Joshi, J. (2001). Computational flow modelling and design of bubble column reactors. *Chemical Engineering Science, 56*(21-22), 5893-5933.

Kazemipoor, M., Hajifaraji, M., Shamshirband, S., Petković, D., & Kiah, M. L. M. (2015). Appraisal of adaptive neuro-fuzzy computing technique for estimating anti-obesity properties of a medicinal plant. *Computer methods and programs in biomedicine, 118*(1), 69-76.

Krishna, R., Urseanu, M., Van Baten, J., & Ellenberger, J. (1999). Influence of scale on the hydrodynamics of bubble columns operating in the churn-turbulent regime: experiments vs. Eulerian simulations. *Chemical Engineering Science, 54*(21), 4903-4911.

Krishna, R., & Van Baten, J. (2003). Mass transfer in bubble columns. *Catalysis today, 79*, 67-75.

Kumar, A., Degaleesan, T., Laddha, G., & Hoelscher, H. (1976). Bubble swarm characteristics in bubble columns. *The Canadian Journal of Chemical Engineering, 54*(5), 503-508.

Laborde-Boutet, C., Larachi, F., Dromard, N., Delsart, O., & Schweich, D. (2009). CFD simulation of bubble column flows: Investigations on turbulence models in RANS approach. *Chemical Engineering Science, 64*(21), 4399-4413.

Lapin, A., Paaschen, T., Junghans, K., & Lübbert, A. (2002). Bubble column fluid dynamics, flow structures in slender columns with large-diameter ring-spargers. *Chemical Engineering Science, 57*(8), 1419-1424.

Lefebvre, S., & Guy, C. (1999). Characterization of bubble column hydrodynamics with local measurements. *Chemical Engineering Science, 54*(21), 4895-4902.

Lei, Y., He, Z., Zi, Y., & Hu, Q. (2007). Fault diagnosis of rotating machinery based on multiple ANFIS combination with GAs. *Mechanical systems and signal processing, 21*(5), 2280-2294.

Li, H., & Prakash, A. (1999). Analysis of bubble dynamics and local hydrodynamics based on instantaneous heat transfer measurements in a slurry bubble column. *Chemical Engineering Science, 54*(21), 5265-5271.

Li, H., & Prakash, A. (2000). Influence of slurry concentrations on bubble population and their rise velocities in a three-phase slurry bubble column. *Powder Technology, 113*(1-2), 158-167.

Li, H., & Prakash, A. (2001). Survey of heat transfer mechanisms in a slurry bubble column. *The Canadian Journal of Chemical Engineering, 79*(5), 717-725.

Li, H., & Prakash, A. (2002). Analysis of flow patterns in bubble and slurry bubble columns based on local heat transfer measurements. *Chemical Engineering Journal, 86*(3), 269-276.

Lin, T.-J., & Wang, S.-P. (2001). Effects of macroscopic hydrodynamics on heat transfer in bubble columns. *Chemical Engineering Science, 56*(3), 1143-1149.

Liu, Y., & Hinrichsen, O. (2014). Study on CFD–PBM turbulence closures based on k–ε and Reynolds stress models for heterogeneous bubble column flows. *Computers & Fluids, 105*, 91-100.

Lopez de Bertodano, M., Lahey Jr, R., & Jones, O. (1994). Turbulent bubbly two-phase flow data in a triangular duct. *Nuclear engineering and design, 146*(1), 43-52.

Luo, X., Lee, D., Lau, R., Yang, G., & Fan, L. S. (1999). Maximum stable bubble size and gas holdup in high-pressure slurry bubble columns. *AIChE Journal, 45*(4), 665-680.

Maalej, S., Benadda, B., & Otterbein, M. (2003). Interfacial area and volumetric mass transfer coefficient in a bubble reactor at elevated pressures. *Chemical Engineering Science, 58*(11), 2365-2376.

Masood, R., & Delgado, A. (2014). Numerical investigation of the interphase forces and turbulence closure in 3D square bubble columns. *Chemical Engineering Science, 108*, 154-168.

Masood, R., Khalid, Y., & Delgado, A. (2015). Scale adaptive simulation of bubble column flows. *Chemical Engineering Journal, 262*, 1126-1136.

McClure, D. D., Aboudha, N., Kavanagh, J. M., Fletcher, D. F., & Barton, G. W. (2015). Mixing in bubble column reactors: Experimental study and CFD modeling. *Chemical Engineering Journal, 264*, 291-301.

McClure, D. D., Kavanagh, J. M., Fletcher, D. F., & Barton, G. W. (2013). Development of a CFD model of bubble column bioreactors: part one–a detailed experimental study. *Chemical Engineering & Technology, 36*(12), 2065-2070.

McClure, D. D., Kavanagh, J. M., Fletcher, D. F., & Barton, G. W. (2014). Development of a CFD model of bubble column bioreactors: part two–comparison of experimental data and CFD predictions. *Chemical Engineering & Technology, 37*(1), 131-140.

McClure, D. D., Norris, H., Kavanagh, J. M., Fletcher, D. F., & Barton, G. W. (2015). Towards a CFD model of bubble columns containing significant surfactant levels. *Chemical Engineering Science, 127*, 189-201.

Michele, V., & Hempel, D. C. (2002). Liquid flow and phase holdup—measurement and CFD modeling for two- and three-phase bubble columns. *Chemical Engineering Science, 57*(11), 1899-1908.

Moazenzadeh, R., Mohammadi, B., Shamshirband, S., & Chau, K.-w. (2018). Coupling a firefly algorithm with support vector regression to predict evaporation in northern Iran. *Engineering Applications of Computational Fluid Mechanics, 12*(1), 584-597.

Mosavi, A., Ozturk, P. and Chau, K.W., 2018. Flood prediction using machine learning models: Literature review. Water, 10(11), p.1536.

Mosavi, A., Rabczuk, T. and Varkonyi-Koczy, A.R., 2017, September. Reviewing the novel machine learning tools for materials design. In International Conference on Global Research and Education (pp. 50-58). Springer, Cham.

Mosavi, A., Salimi, M., Faizollahzadeh Ardabili, S., Rabczuk, T., Shamshirband, S. and Varkonyi-Koczy, A.R., 2019. State of the art of machine learning models in energy systems, a systematic review. Energies, 12(7), p.1301.

Pfleger, D., & Becker, S. (2001). Modelling and simulation of the dynamic flow behaviour in a bubble column. *Chemical Engineering Science, 56*(4), 1737-1747.

Pino, L., Solari, R., Siquier, S., Antonio Estevez, L., Yepez, M., & Saez, A. (1992). Effect of operating conditions on gas holdup in slurry bubble columns with a foaming liquid. *Chemical Engineering Communications, 117*(1), 367-382.

Pourtousi, M. (2012). *Simulation of Particle Motion in Incompressible Fluid by Lattice Boltzmann MRT Model.* Universiti Teknologi Malaysia.

Pourtousi, M. (2016). *CFD modelling and anfis development for the hydrodynamics prediction of bubble column reactor ring sparger.* University of Malaya.

Pourtousi, M., Ganesan, P., Kazemzadeh, A., Sandaran, S. C., & Sahu, J. (2015). Methane bubble formation and dynamics in a rectangular bubble column: A CFD study. *Chemometrics and Intelligent Laboratory Systems, 147*, 111-120.

Pourtousi, M., Ganesan, P., & Sahu, J. (2015). Effect of bubble diameter size on prediction of flow pattern in Euler–Euler simulation of homogeneous bubble column regime. *Measurement, 76*, 255-270.

Pourtousi, M., Ganesan, P., Sandaran, S. C., & Sahu, J. (2016). Effect of ring sparger diameters on hydrodynamics in bubble column: A numerical investigation. *Journal of the Taiwan Institute of Chemical Engineers, 69*, 14-24.

Pourtousi, M., Sahu, J., Ganesan, P., Shamshirband, S., & Redzwan, G. (2015). A combination of computational fluid dynamics (CFD) and adaptive neuro-fuzzy system (ANFIS) for prediction of the bubble column hydrodynamics. *Powder Technology, 274*, 466-481.

Pourtousi, M., Sahu, J. N., & Ganesan, P. (2014). Effect of interfacial forces and turbulence models on predicting flow pattern inside the bubble column. *Chemical Engineering and Processing: Process Intensification, 75*, 38-47. doi:10.1016/j.cep.2013.11.001

Pourtousi, M., Sahu, J. N., Ganesan, P., Shamshirband, S., & Redzwan, G. (2015). A combination of computational fluid dynamics (CFD) and adaptive neuro-fuzzy system (ANFIS) for prediction of the bubble column hydrodynamics. *Powder Technology, 274*, 466-481. doi:10.1016/j.powtec.2015.01.038

Pourtousi, M., Zeinali, M., Ganesan, P., & Sahu, J. N. (2015). Prediction of multiphase flow pattern inside a 3D bubble column reactor using a combination of CFD and ANFIS. *RSC Advances, 5*(104), 85652-85672. doi:10.1039/c5ra11583c

Prakash, A., Margaritis, A., Li, H., & Bergougnou, M. (2001). Hydrodynamics and local heat transfer measurements in a bubble column with suspension of yeast. *Biochemical Engineering Journal, 9*(2), 155-163.

Rabha, S., Schubert, M., & Hampel, U. (2013). Intrinsic flow behavior in a slurry bubble column: a study on the effect of particle size. *Chemical Engineering Science, 93*, 401-411.

Rampure, M. R., Kulkarni, A. A., & Ranade, V. V. (2007). Hydrodynamics of bubble column reactors at high gas velocity: experiments and computational fluid dynamics (CFD) simulations. *Industrial & Engineering Chemistry Research, 46*(25), 8431-8447.

Riahi-Madvar, H., Dehghani, M., Seifi, A., Salwana, E., Shamshirband, S., Mosavi, A. and Chau, K.W., 2019. Comparative analysis of soft computing techniques RBF, MLP, and ANFIS with MLR and MNLR for predicting grade-control scour hole geometry. Engineering Applications of Computational Fluid Mechanics, 13(1), pp.529-550.

Ruzicka, M., Zahradnık, J., Drahoš, J., & Thomas, N. (2001). Homogeneous–heterogeneous regime transition in bubble columns. *Chemical Engineering Science, 56*(15), 4609-4626.

Ryoo, J., Dragojlovic, Z., & Kaminski, D. A. (2005). Control of convergence in a computational fluid dynamics simulation using ANFIS. *IEEE Transactions on fuzzy systems, 13*(1), 42-47.

Rzehak, R., & Krepper, E. (2013). CFD modeling of bubble-induced turbulence. *International Journal of Multiphase Flow, 55*, 138-155.

Rezakazemi, M., Mosavi, A. and Shirazian, S., 2019. ANFIS pattern for molecular membranes separation optimization. Journal of Molecular Liquids, 274, pp.470-476.

Şal, S., Gül, Ö. F., & Özdemir, M. (2013). The effect of sparger geometry on gas holdup and regime transition points in a bubble column equipped with perforated plate spargers. *Chemical Engineering and Processing: Process Intensification, 70*, 259-266.

Sanyal, J., Marchisio, D. L., Fox, R. O., & Dhanasekharan, K. (2005). On the comparison between population balance models for CFD simulation of bubble columns. *Industrial & Engineering Chemistry Research, 44*(14), 5063-5072.

Sanyal, J., Vásquez, S., Roy, S., & Dudukovic, M. (1999). Numerical simulation of gas–liquid dynamics in cylindrical bubble column reactors. *Chemical Engineering Science, 54*(21), 5071-5083.

Schäfer, R., Merten, C., & Eigenberger, G. (2002). Bubble size distributions in a bubble column reactor under industrial conditions. *Experimental thermal and fluid science, 26*(6-7), 595-604.

Schumpe, A., & Grund, G. J. T. C. J. o. C. E. (1986). The gas disengagement technique for studying gas holdup structure in bubble columns. *64*(6), 891-896.

Schurter, K. C., & Roschke, P. N. (2000). *Fuzzy modeling of a magnetorheological damper using ANFIS.* Paper presented at the Fuzzy Systems, 2000. FUZZ IEEE 2000. The Ninth IEEE International Conference on.

Shah, Y., Kelkar, B. G., Godbole, S., & Deckwer, W. D. (1982). Design parameters estimations for bubble column reactors. *AIChE Journal, 28*(3), 353-379.

Shamshirband, S., Babanezhad, M., & Mosavi, A. (2019). Prediction of Flow Characteristics in the Bubble Column Reactor by the Artificial Pheromone-Based Communication of Biological Ants. Preprints 2019, 2019050025 (doi: 10.20944/preprints201905.0025.v1).

Shimizu, K., Takada, S., Minekawa, K., & Kawase, Y. (2000). Phenomenological model for bubble column reactors: prediction of gas hold-ups and volumetric mass transfer coefficients. *Chemical Engineering Journal, 78*(1), 21-28.

Silva, M. K., d'Ávila, M. A., & Mori, M. (2012). Study of the interfacial forces and turbulence models in a bubble column. *Computers & Chemical Engineering, 44*, 34-44.

Simonnet, M., Gentric, C., Olmos, E., & Midoux, N. (2007). Experimental determination of the drag coefficient in a swarm of bubbles. *Chemical Engineering Science, 62*(3), 858-866.

Simonnet, M., Gentric, C., Olmos, E., & Midoux, N. (2008). CFD simulation of the flow field in a bubble column reactor: Importance of the drag force formulation to describe regime transitions. *Chemical Engineering and Processing: Process Intensification, 47*(9-10), 1726-1737.

Sobrino, C., Acosta-Iborra, A., Izquierdo-Barrientos, M. A., & De Vega, M. (2015). Three-dimensional two-fluid modeling of a cylindrical fluidized bed and validation of the maximum entropy method to determine bubble properties. *Chemical Engineering Journal, 262*, 628-639.

Sokolichin, A., & Eigenberger, G. (1994). Gas—liquid flow in bubble columns and loop reactors: Part I. Detailed modelling and numerical simulation. *Chemical Engineering Science, 49*(24), 5735-5746.

Sokolichin, A., Eigenberger, G., & Lapin, A. (2004). Simulation of buoyancy driven bubbly flow: established simplifications and open questions. *AIChE Journal, 50*(1), 24-45.

Tabib, M. V., Roy, S. A., & Joshi, J. B. (2008). CFD simulation of bubble column—an analysis of interphase forces and turbulence models. *Chemical Engineering Journal, 139*(3), 589-614.

Taha, M. R., Noureldin, A., & El-Sheimy, N. (2001). *Improving INS/GPS positioning accuracy during GPS outages using fuzzy logic.* Paper presented at the Proceedings of the 16th International Technical Meeting of the Satellite Division of The Institute of Navigation (ION GPS/GNSS 2003).

Takagi, T., & Sugeno, M. (1985). Fuzzy identification of systems and its applications to modeling and control. *IEEE transactions on systems, man, and cybernetics*(1), 116-132.

Tang, C., & Heindel, T. J. (2004). Time-dependent gas holdup variation in an air–water bubble column. *Chemical Engineering Science, 59*(3), 623-632.

Thorat, B., & Joshi, J. (2004). Regime transition in bubble columns: experimental and predictions. *Experimental thermal and fluid science, 28*(5), 423-430.

Thorat, B., Joshi, J. J. E. T., & Science, F. (2004). Regime transition in bubble columns: experimental and predictions. *28*(5), 423-430.

Van Baten, J., Ellenberger, J., & Krishna, R. (2003). Hydrodynamics of internal air-lift reactors: experiments versus CFD simulations. *Chemical Engineering and Processing: Process Intensification, 42*(10), 733-742.

Vandu, C., & Krishna, R. (2004). Volumetric mass transfer coefficients in slurry bubble columns operating in the churn-turbulent flow regime. *Chemical Engineering and Processing: Process Intensification, 43*(8), 987-995.

Vargas, R., Mosavi, A. and Ruiz, R., 2017. Deep learning: a review. Advances in Intelligent Systems and Computing. Springer 29 (8). 232-244.

Varol, Y., Koca, A., Oztop, H. F., & Avci, E. (2008). Analysis of adaptive-network-based fuzzy inference system (ANFIS) to estimate buoyancy-induced flow field in partially heated triangular enclosures. *Expert Systems with Applications, 35*(4), 1989-1997.

Veera, U. P., Kataria, K., & Joshi, J. (2004). Effect of superficial gas velocity on gas hold-up profiles in foaming liquids in bubble column reactors. *Chemical Engineering Journal, 99*(1), 53-58.

Verma, A., & Rai, S. (2003). Studies on surface to bulk ionic mass transfer in bubble column. *Chemical Engineering Journal, 94*(1), 67-72.

Wang, H., Jia, X., Wang, X., Zhou, Z., Wen, J., & Zhang, J. (2014). CFD modeling of hydrodynamic characteristics of a gas–liquid two-phase stirred tank. *Applied Mathematical Modelling, 38*(1), 63-92.

Wang, S., Arimatsu, Y., Koumatsu, K., Furumoto, K., Yoshimoto, M., Fukunaga, K., & Nakao, K. (2003). Gas holdup, liquid circulating velocity and mass transfer properties in a mini-scale external loop airlift bubble column. *Chemical Engineering Science, 58*(15), 3353-3360.

Wu, C., & Chau, K. (2011). Rainfall–runoff modeling using artificial neural network coupled with singular spectrum analysis. *Journal of Hydrology, 399*(3-4), 394-409.

Xiao, Q., Yang, N., & Li, J. (2013). Stability-constrained multi-fluid CFD models for gas–liquid flow in bubble columns. *Chemical Engineering Science, 100*, 279-292.

Xing, C., Wang, T., & Wang, J. (2013). Experimental study and numerical simulation with a coupled CFD–PBM model of the effect of liquid viscosity in a bubble column. *Chemical Engineering Science, 95*, 313-322.

Yaseen, Z. M., Sulaiman, S. O., Deo, R. C., & Chau, K.-W. (2018). An enhanced extreme learning machine model for river flow forecasting: State-of-the-art, practical applications in water resource engineering area and future research direction. *Journal of Hydrology*.

Ziegenhein, T., Rzehak, R., Krepper, E., & Lucas, D. (2013). Numerical Simulation of Polydispersed Flow in Bubble Columns with the Inhomogeneous Multi-Size-Group Model. *Chemie Ingenieur Technik, 85*(7), 1080-1091.

Ziegenhein, T., Rzehak, R., & Lucas, D. (2015). Transient simulation for large scale flow in bubble columns. *Chemical Engineering Science, 122*, 1-13. doi:10.1016/j.ces.2014.09.022